\documentclass[10pt,twocolumn,letterpaper]{article}
\usepackage[table]{xcolor}

\usepackage[pagenumbers]{iccv} 
\usepackage{amsmath,amsfonts,bm}
\def\eqref#1{equation~\ref{#1}}
\def\1{\bm{1}}

\def\rvx{{\mathbf{x}}}

\DeclareMathAlphabet{\mathsfit}{\encodingdefault}{\sfdefault}{m}{sl}
\SetMathAlphabet{\mathsfit}{bold}{\encodingdefault}{\sfdefault}{bx}{n}

\newcommand{\E}{\mathbb{E}}

\usepackage{bbding}
\usepackage{pifont}
\usepackage{algorithm}  % 提供algorithm环境
\usepackage{algpseudocode}  % 提供算法伪代码命令

\definecolor{iccvblue}{rgb}{0.21,0.49,0.74}
\usepackage[pagebackref,breaklinks,colorlinks,allcolors=iccvblue]{hyperref}
\newcommand{\spara}[1]{\noindent\textbf{#1.}}
\usepackage{multirow}
\usepackage{media9} % 插入视频
\usepackage{subcaption}

\newcommand{\name}{Trajectory Distribution Matching\xspace}
\newcommand{\namebf}{\textbf{T}rajectory \textbf{D}istribution \textbf{M}atching\xspace}
\newcommand{\shortname}{TDM\xspace}

\title{Learning Few-Step Diffusion Models by \name}

\author{Yihong Luo $^{1}$\protect\footnotemark[1]  \hspace{5mm} 
Tianyang Hu $^{2}$ \hspace{5mm}
Jiacheng Sun $^{2}$ \hspace{5mm} 
Yujun Cai $^{3}$ \hspace{5mm} 
Jing Tang $^{4,1}$\protect\footnotemark[2] \hspace{5mm}  \\ \vspace{-3mm} \\
$^1$ HKUST \hspace{5mm} $^2$ Huawei Noah’s Ark Lab \hspace{5mm} $^3$ NUS \hspace{5mm}$^4$ HKUST (GZ) 
}

\begin{document}

\twocolumn[{%
 \renewcommand\twocolumn[1][]{#1}%
 \maketitle
 \vspace{-6mm}
 \centering
 \includegraphics[width=\textwidth]{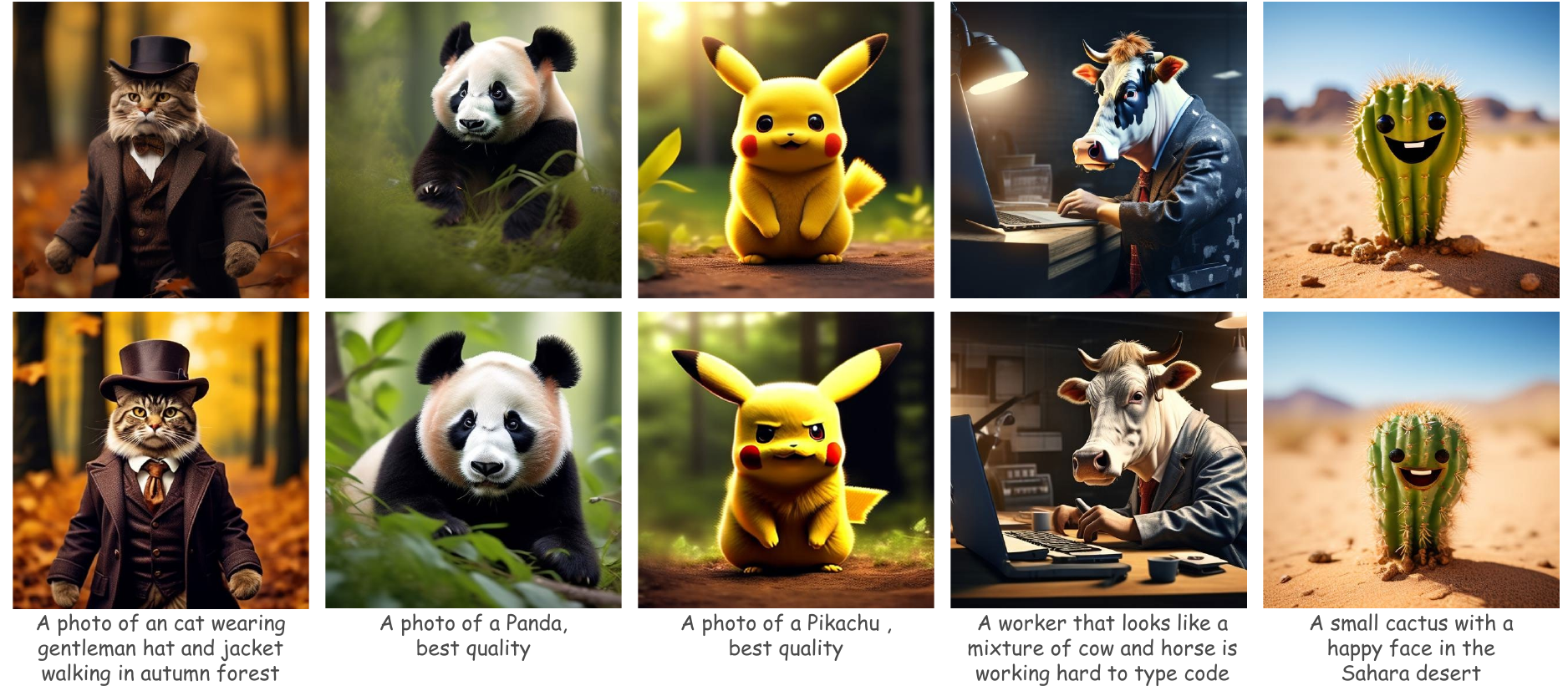}
 \vspace{-8mm}
 \captionof{figure}{ \textbf{User Study Time!} Which one do you think is better? Some images are generated by \textbf{Pixart-$\alpha$ (50 NFE)}. Some images are generated by \textbf{\shortname (4 NFE)}, distilling from Pixart-$\alpha$ in a \textit{data-free way} with merely \textbf{500 training iterations} and \textbf{2 A800 hours}. All images are generated from the same initial noise. We put the location of generated images by \shortname in footnote\protect\footnotemark[3].
    \label{fig:teaser}
   }
   \vspace{2mm}
}]
% \maketitle
\footnotetext[1]{Work was partly done during an internship at Huawei Noah’s Ark Lab.}
\footnotetext[2]{Corresponding Author.}
\footnotetext[3]{\textbf{\shortname (left to right)}:
bottom, bottom, top, bottom, top.}
\begin{abstract}
Accelerating diffusion model sampling is crucial for efficient AIGC deployment. 
While diffusion distillation methods---based on distribution matching \citep{yin2023one,luo2023diff} and trajectory matching \citep{salimans2021progressive,song2023consistency}---reduce sampling to as few as one step, they fall short on complex tasks like text-to-image generation. 
Few-step generation offers a better balance between speed and quality, but existing approaches face a persistent trade-off: distribution matching lacks flexibility for multi-step sampling, while trajectory matching often yields suboptimal image quality.
To bridge this gap, we propose learning few-step diffusion models by \namebf (\textbf{\shortname}), a unified distillation paradigm that combines the strengths of distribution and trajectory matching. 
Our method introduces a data-free score distillation objective, aligning the student’s trajectory with the teacher’s at the distribution level. 
Further, we develop a sampling-steps-aware objective that decouples learning targets across different steps, enabling more adjustable sampling.
This approach supports both deterministic sampling for superior image quality and flexible multi-step adaptation, achieving state-of-the-art performance with remarkable efficiency. 
Our model, \shortname, outperforms existing methods on various backbones, such as SDXL and PixArt-$\alpha$, delivering superior quality and significantly reduced training costs.
In particular, our method distills PixArt-$\alpha$ into a 4-step generator that outperforms its teacher on real user preference at 1024 resolution. This is accomplished with 500 iterations and 2 A800 hours---a mere 0.01\% of the teacher's training cost. 
In addition, our proposed \shortname can be extended to accelerate text-to-video diffusion. Notably, \shortname can outperform its teacher model (CogVideoX-2B) by using only 4 NFE on VBench, improving the total score from 80.91 to 81.65. Project page: \url{https://tdm-t2x.github.io/}.
\end{abstract}

\vspace{-5mm}
\section{Introduction}
\vspace{-3mm}
\label{sec:intro}

\begin{figure*}[!t]
    \centering
    \includegraphics[width=1\linewidth]{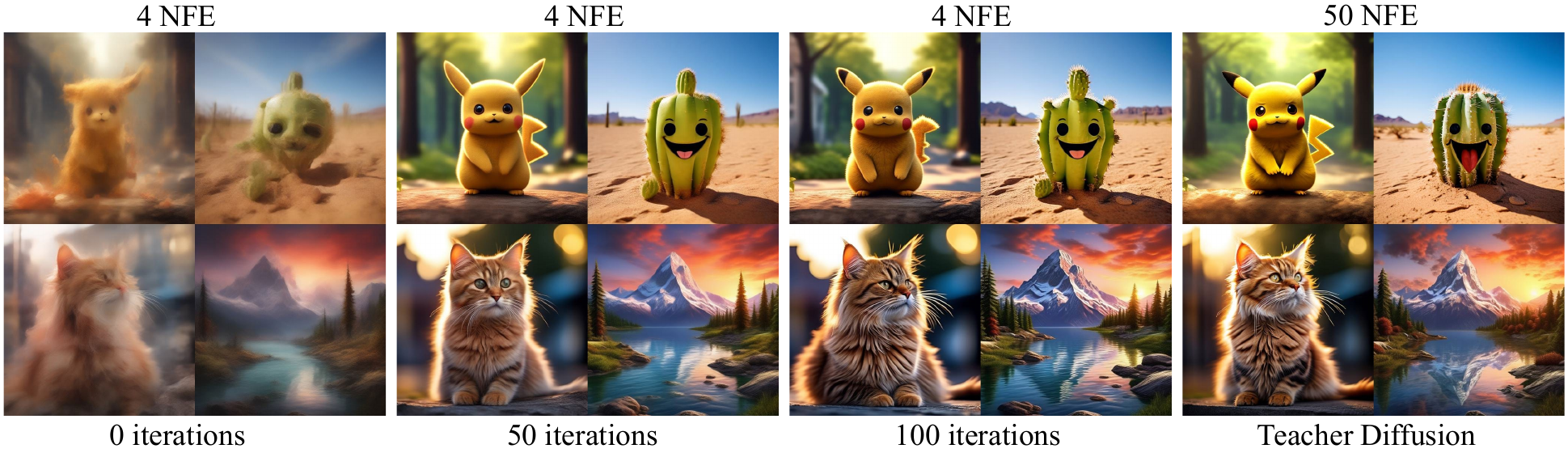}
    \vspace{-8mm}
    \caption{ The comparison between \textbf{Four-step} generated images by \textbf{\shortname} under different training iterations and pre-trained diffusion models with 25 steps and 5.5 CFG. It can be seen that the ultra-fast convergence of our method, without sacrificing the sample quality.}
    \label{fig:ultra_fast_convergence}
    \vspace{-6mm}
\end{figure*}

    Accelerating diffusion model sampling is essential for the efficient deployment of AIGC models. 
    Thanks to various diffusion distillation techniques, the number of sampling steps has been drastically reduced to as few as one. 
    Most notably, approaches based on distribution matching through score distillation, such as DMD~\cite{yin2023one}, Diff-Instruct~\cite{luo2023diff}, and SiD~\cite{sid}, have demonstrated strong performance in this regime. However, even state-of-the-art one-step generation methods fall short in complex tasks, particularly in text-to-image generation.
    In such cases, \textbf{few-step generation} strikes a better balance between speed and image quality.

    Yet, existing distribution matching methods are primarily designed for one-step generation, optimizing solely for performance at this extreme and lacking the flexibility to extract extra information with increased sampling steps.
    An alternative approach, commonly known as trajectory distillation~\cite{salimans2021progressive,luhman2021knowledge,song2023consistency}, operates on the instance level, aiming to simulate the original diffusion generation trajectories in as few steps as possible. 
    However, matching entire instance-level trajectories with fewer steps poses substantial challenges, requiring high model capacity. Consequently, its performance in few-step generation remains suboptimal.

    To bridge this gap, we propose a novel framework that non-trivially unifies trajectory distillation and distribution matching, termed \namebf (\textbf{\shortname}).  
     Specifically, our approach aligns the student’s trajectory with the teacher’s at the distribution level using a novel, data-free score distillation objective. 
      \shortname supports deterministic sampling which inherently has a better few-step performance than stochastic sampling, enabling faster convergence \citep{lu2022dpm, lu2023dpm, bao2022analytic,ddim}.
     Our method combines the strengths of both distribution and trajectory matching.
     On the distribution matching side, the addition of trajectory information enables us to better inherit teachers' knowledge in a fine-grained fashion for multi-step generation.
     On the trajectory matching side, since we match the pre-trained diffusion distribution at different timesteps, the learning of the generator becomes easier, and better performance is expected from the same model capacity. 

    To accommodate diverse use cases, an ideal generator should also offer users flexible control over the number of sampling steps. 
    However, existing works in text-to-image diffusion distillation either only support less-adjustable steps based on deterministic sampling~\cite{salimans2021progressive,yan2024perflow,sdxllight} or adjustable steps based on stochastic sampling, which sacrifices image quality for flexibility~\cite{luo2023latent,ren2024hypersd}. 
    To address these limitations, we propose a novel sampling-steps-aware objective that decouples learning targets among different sampling steps, enabling flexible deterministic sampling. We name the model trained with this technique as \shortname-unify.
    Furthermore, inspired by our new perspective built on the commonalities between \shortname and CMs~\cite{song2023consistency}, we introduce a surrogate training objective that employs the Pseudo-Huber metric, which has proven effective in iCT~\cite{song2024improved}.

    Our method achieves state-of-the-art (SOTA) performance across various text-to-image backbones with remarkable efficiency. 
    On SD-v1.5~\cite{rombach2022high}, our \shortname-unify demonstrates superior performance in both 1-step and 4-step generation, substantially outperforming existing approaches. On distilling SDXL~\cite{podell2023sdxl} for 4-step inference, our method surpasses SDXL-Lightning~\cite{sdxllight} by +2.17 in Human Preference Score (HPS)~\citep{wu2023human} and +1.05 in CLIP Score, outperforms Hyper-SD~\cite{ren2024hypersd} by +0.74 in HPS and +1.81 in CLIP Score, and exceeds DMD2~\cite{dmd2} by +3.42 in HPS and +0.57 in CLIP Score. Notably, our approach demonstrates exceptional training efficiency, requiring only 2 A800 days for SDXL compared to LCM's~\cite{luo2023latent} 32 A100 days and DMD2's 160 A100 days. On distilling PixArt-$\alpha$~\cite{chen2024pixartalpha} for 4-step inference, our method surpasses the multi-step teacher in terms of HPS, AeS, and user preference. This is achieved by merely 500 training iterations and 2 A800 hours, only 0.01\% of the teacher's training cost. Additionally, our method can distill the video diffusion models to a 4-step generator with better performance. In particular, we distill the CogVideoX-2B~\cite{yang2024cogvideox} to a 4-step generator by \shortname, improving the Vbench~\cite{huang2023vbench} total score from 80.91 to 81.65.

\begin{figure*}
    \centering
    \includegraphics[width=1\linewidth]{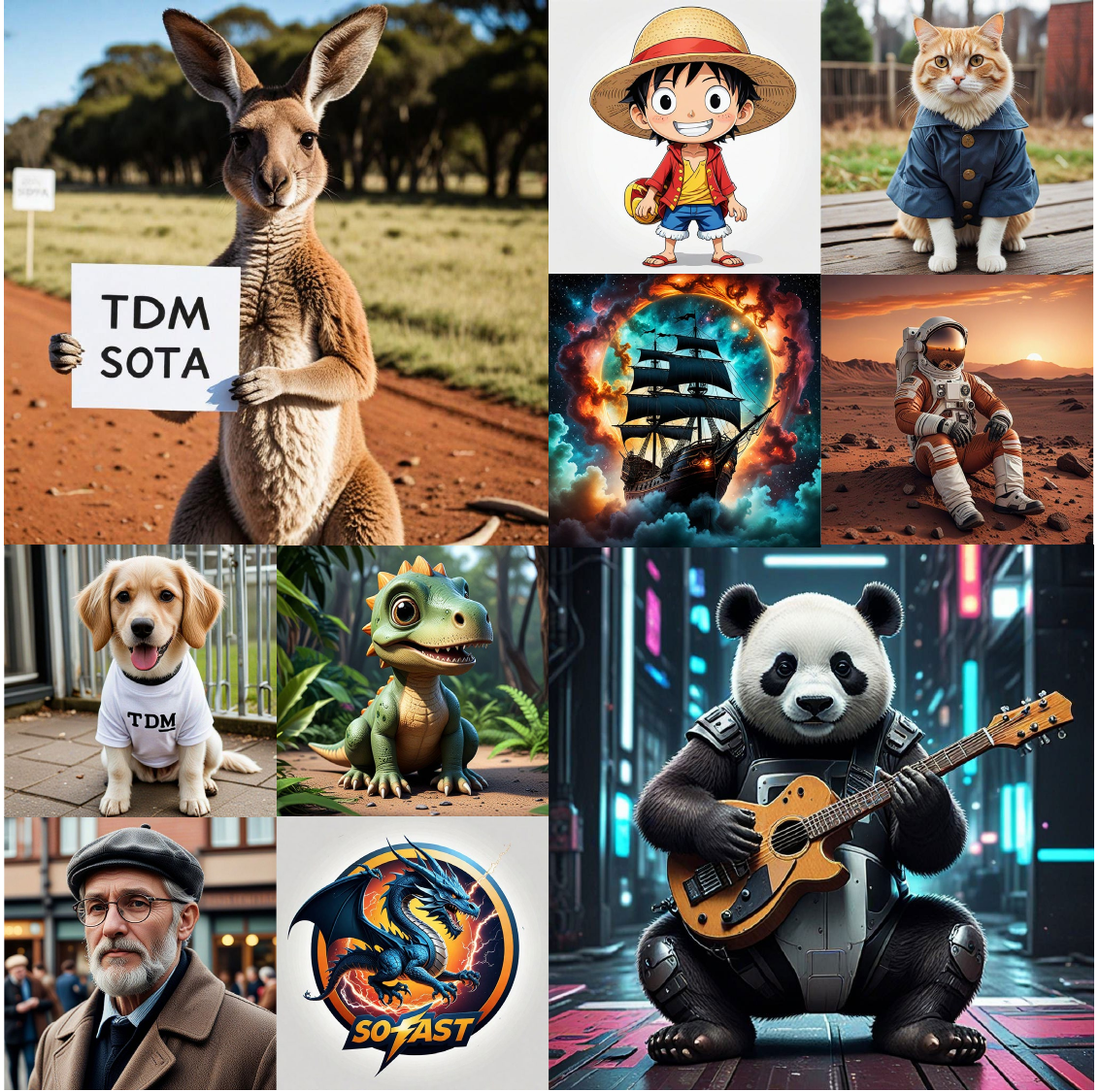}
    \vspace{-4mm}
    \caption{Additional Samples by TDM with 4-step generation on SDXL backbone.}
    \label{fig:add_sdxl}
    \vspace{-4mm}
\end{figure*}
\vspace{-3mm}
\section{Background}
\vspace{-2mm}
\spara{Diffusion Models (DMs)} 
DMs~\citep{sohl2015deep, ho2020denoising} define a forward diffusion process that gradually injects Gaussian noises into the data in $T$ steps: $q(\rvx_t|\rvx) \triangleq \mathcal{N}(\rvx_t; \rvx, \sigma_t^2\textbf{I}),$
where $\sigma_t$ specifies the noise schedule. 
The diffused samples can be directly obtained by $\rvx_t = \rvx + \sigma_t \epsilon$, where $\epsilon \sim \mathcal{N}(\mathbf{0},\mathbf{I})$. 
Then the diffusion network $f_\phi$ is trained by denoising via: $\E_{t,\rvx,\epsilon} \omega_t || f_\phi(\rvx_t,t) - \rvx||_2^2,$
where $\omega_t$ controls the importance of the denoising at different timesteps.
After training, we can estimate the score of diffused samples via:
\begin{equation}
\small
    \nabla_{\rvx_t} \log p_t(\rvx_t) \approx  s_\phi(\rvx_t,t) = - \tfrac{\rvx_t - f_\phi(\rvx_t,t)}{\sigma_t^2},
\end{equation}
where $p_t(\rvx_t) = \int q(\rvx_t|\rvx)p_d(\rvx)d\rvx$.
Sampling from DMs can be achieved by solving corresponding Probability Flow ODE (PF-ODE) via multiple steps (typically $\geq$ 25)~\cite{song2020score}.

\spara{Diffusion Distillation} Recently, there are two appealing ways to explicitly transfer knowledge from pre-trained DMs to few-step students: 1) Trajectory distillation~\cite{luhman2021knowledge,on_distill,luhman2021knowledge,salimans2021progressive,song2023consistency,song2024improved,yan2024perflow}, which typically aims to simulate teacher ODE trajectories on the instance level. These methods suffer from difficult instance-level matching and numerical errors when solving teacher's PF-ODE;
2) Distribution matching via score distillation~\cite{yin2023one,luo2023diff,sid}, which distills from pre-trained DMs on distribution level via a certain distribution divergence. These methods mostly ignore the intermediate steps of trajectory, making it hard to extract extra information with increased sampling steps. More discussion on related works deferred to \cref{sec:related_work} due to limited space.

\vspace{-2mm}
\section{Methodology}
\vspace{-1mm}
Despite the success of existing diffusion distillation methods, achieving industry-ready few-step performance remains challenging. The challenge might stem from trajectory distillation and distribution matching through score distillation both of which suffer from different limitations as we discussed above.
To address these limitations, we propose \name (\shortname), a novel framework that unifies trajectory distillation and distribution matching, aligning the student's trajectory with the teacher's at the distribution level. 
First, we introduce a distribution-level trajectory matching objective that enables efficient knowledge transfer while avoiding numerical errors in solving teacher's PF-ODE (\cref{sec:main_frame}). Second, we propose a sampling-steps-aware objective that supports flexible step adjustment (\cref{sec:flexible_steps}). Finally, we develop a surrogate training objective inspired by consistency models to enhance optimization 
 (\cref{sec:surrogate_obj}).

\subsection{\name}
\label{sec:main_frame}
The diffusion distillation via simulating the teacher trajectory typically minimizes the following objective~\cite{luhman2021knowledge,salimans2021progressive}:
\begin{equation}
\small
    \min_\theta d(\mathrm{ODESolver}(\rvx_t, f_\phi, t, s), \mathrm{DDIM}(\rvx_t, f_\theta, t, s)),
\end{equation}
where $d(\cdot,\cdot)$ is a distance metric, 
$\rvx_t$ is the diffused samples,
$f_\phi$ is the teacher network, and $f_\theta$ is the student network. 
The learning objective can be re-written as the following joint forward KL divergence:
\begin{equation}
\small
\begin{aligned}
    \min_\theta \mathrm{KL}(p_t(\rvx_t)p_\phi(\rvx_s|\rvx_t) || p_t(\rvx_t) p_\theta(\rvx_s|\rvx_t)) \\ 
\end{aligned}
\end{equation}
where $p_t(\rvx_t)$ is the marginal distribution of diffused samples, $p_\phi(\rvx_s|\rvx_t)$ is the backward diffusion process which can be sampled by ODESolver, and $p_\theta(\rvx_s|\rvx_t)$ is the desired one-step denoising distribution.

The objective suffers from multiple issues: 1) This objective enforces a point-to-point match on the trajectory which makes it difficult to achieve optimal performance due to limited model capacity; 2) This objective requires solving ODE via multi-step DMs which is not only computationally expensive but also potentially suffers form numerical error.

We argue that simulating the teacher's ODE trajectories can maximize the inheritance of teacher knowledge, thereby reducing learning difficulty, accelerating convergence, and improving distillation performance. However, existing trajectory distillation methods have suffered severe performance degradation due to the aforementioned issues.
 
To address these issues, we propose to align the ODE trajectories of the distillation model with the teacher model at corresponding timesteps at the distribution level.
Specifically, at each training iteration, we use an ODESolver (e.g., DPMSolver~\cite{dpmsolver}) to sample from the K-step student model $p_\theta$ and save the intermediate samples $\rvx_{t_i}$ on the trajectory. Then motivated by the success of score distillation such as Diff-Instruct~\citep{luo2023diff} and VSD~\cite{wang2023prolificdreamer}, we seek to minimize the following marginal reverse KL divergence:
\begin{equation}
\small
    L(\theta) =  \sum_{i=0}^{K-1}\mathrm{KL}(p_{\theta,t_i}(\rvx_{t_i})||p_{\phi,t_i}(\rvx_{t_i})),
\end{equation}
where $p_\phi$ is the pre-trained teacher DM, $\rvx_{t_i}$ is the samples on the student trajectory at timestep $t_i$, and $p_{\theta,t_i}(\rvx_{t_i})$ denotes the implicit distribution of $\rvx_{t_i}$. 
By default, we let $t_i = \frac{T}{K}i$, where $T$ is the terminal timestep. 

The advantage of the objective is twofold: 1) we only need to sample from the student which is data-free and computationally efficient; 2) we can simulate the teacher trajectory at the distribution level without requiring sampling from the teacher. This also avoids numerical error in solving teacher ODE. 
It is crucial to note that while we need to ``solve" the student ODE, we are essentially just training a K-step generator that is parameterized by the discrete form of the ODE sampler.

Following previous works~\cite{luo2023diff,yin2023one}, we diffuse the distribution to align the training objective with the diffusion process for better transferring knowledge from pre-trained DMs. The final objective becomes:
\begin{equation}
\label{eq:obj}
\small
    L(\theta) = \sum_{i=0}^{K-1}\sum_{\tau = t_i}^{t_{i+1}} \lambda_{\tau}\mathrm{KL}(p_{\theta, \tau|t_i}(\rvx_\tau) || p_{\phi,\tau}(\rvx_{\tau})),
\end{equation}
where $p_{\theta, \tau|t_i}(\rvx_\tau)\triangleq \int q(\rvx_\tau|\rvx_{t_i}) p_{\theta,t_i}(\rvx_{t_i})d\rvx_{t_i}$ denotes a marginal diffused distribution at timestep $\tau$.

\begin{figure}
    \centering
    \includegraphics[width=1\linewidth]{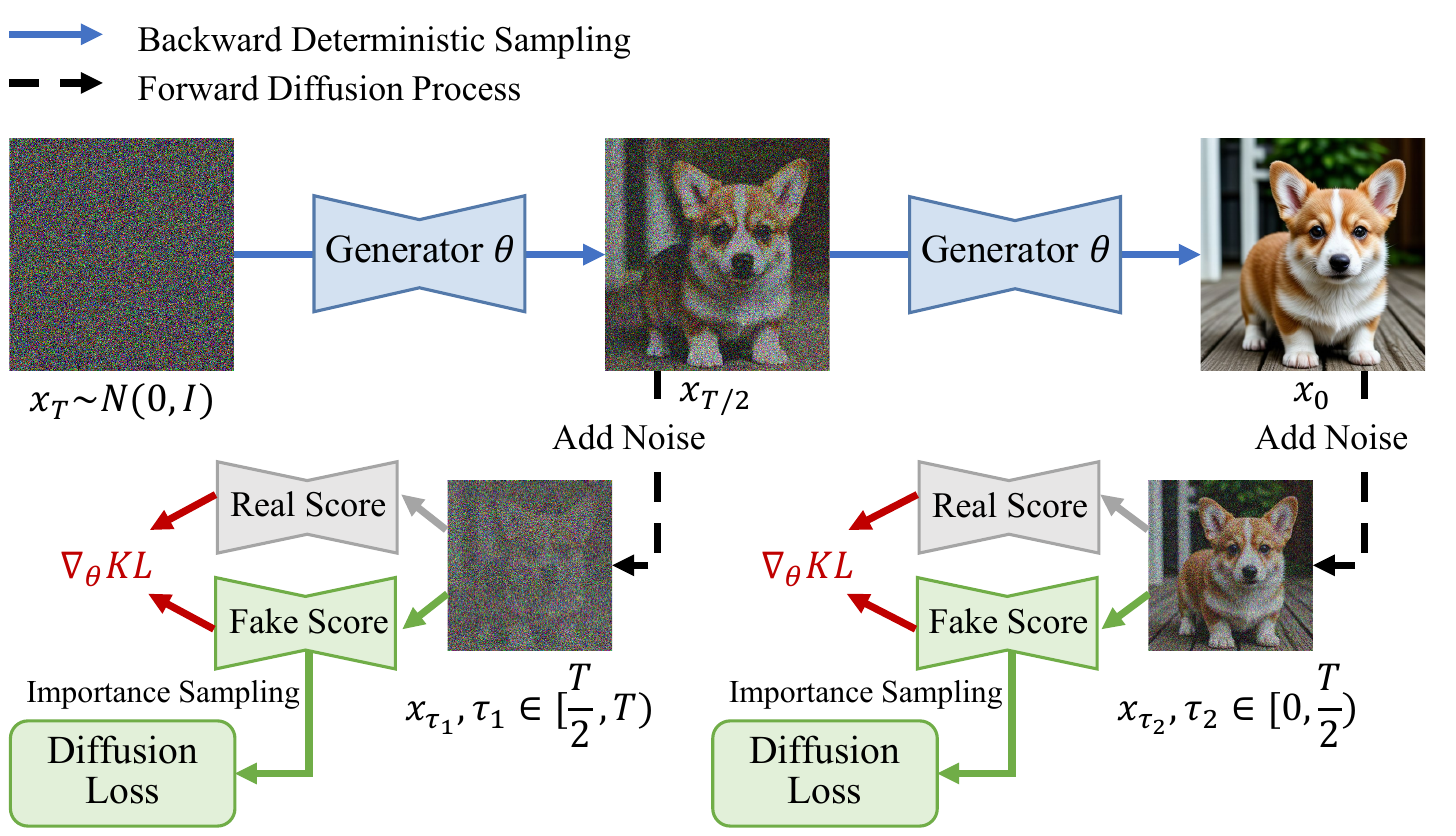}
    \vspace{-4mm}
    \caption{\textbf{\name}. An illustration of training 2-step generator by \shortname in a \textit{data-free} way.}
    \label{fig:framework}
\end{figure}

The gradient of the proposed objective can be computed as follows:
\begin{equation}
\small
% \notag
\label{eq:gradient}
\begin{aligned}
    \nabla_\theta L(\theta) & = \sum_{i=0}^{K-1}\sum_{\tau = t_i}^{t_{i+1}}
   \lambda_{\tau} [\nabla_{\rvx_\tau} \log p_{\theta, \tau|t_i}(\rvx_\tau) - s_\phi(\rvx_\tau,\tau)]\frac{\partial{\rvx_{t_i}}}{\partial \theta}\nonumber\\
    & \approx \sum_{i=0}^{K-1}\sum_{\tau = t_i}^{t_{i+1}}
  \lambda_{\tau} [s_\psi(\rvx_\tau,\tau) - s_\phi(\rvx_\tau,\tau)]\frac{\partial{\rvx_{t_i}}}{\partial \theta}.
\end{aligned}
\end{equation}
Since the score of student distribution is hard to access, we use another DM $s_\psi$ for approximation. Following GAN's tradition~\cite{goodfellow2014generative}, we call $s_\phi$ as real score and $s_\psi$ as fake score. We briefly summarize the learning of generator in \cref{fig:framework}.

For ease of reference, we rewrite the objective for training student $\theta$ with the same gradient in \cref{eq:gradient} as follows:
\begin{equation}
\label{eq:obj_g}
\small
    L(\theta) = \sum_{i=0}^{K-1}\sum_{\tau = t_i}^{t_{i+1}} \lambda_{\tau} || \rvx_{t_i} - \mathrm{sg}(\Tilde{\rvx}_{t_i}) ||_2^2,
\end{equation}
where $\Tilde{\rvx}_{t_i} = \rvx_{t_i} + \lambda_{\tau} [s_\phi(\rvx_\tau,\tau) - s_\psi(\rvx_\tau,\tau)]$, 
$\rvx_{t_i}$ is the sample on the student trajectory, and $\mathrm{sg}$ denotes stop gradient. The $\Tilde{\rvx}_{t_i}$ can be regarded as revised samples obtained by performing one gradient descent step on the reverse KL divergence.

Note that we deliberately ensure that the diffused intervals $[t_i,t_{i+1}]$ do not overlap in \cref{eq:obj,eq:obj_g}. The main reason for this is that trajectory samples at different timesteps have different distributions. Theoretically, we should use different fake scores for modeling. However, non-overlapping intervals allow us to use the same fake score for modeling. In this scenario, the timestep can naturally separate samples from different distributions. Moreover, when backpropagating through $\rvx_t$, we only consider one ODE step for saving GPU memory. 

\noindent\textbf{Learning Fake Score $s_\psi$} 
The fake score can be trained by denoising as follows:
\begin{equation}
\small
    L(\psi) = \sum_{i=0}^{K-1} \E_{p_{\theta,t_i}(\rvx_{t_i})}\E_{q(\rvx_\tau | \hat \rvx_{t_i})} \omega_{\tau}|| f_\psi(\rvx_\tau,\tau) - \hat \rvx_{t_i}||_2^2,
\end{equation}
where $\hat \rvx_{t_i}$ denotes the clean sample corresponding to noisy samples $\rvx_{t_i}$. 
Since training the generator only requires computing the score of noisy samples diffused from $\rvx_{t_i}$, we introduce \textit{importance sampling} to enhance the efficiency of learning fake score:
\begin{equation}
\small
\notag
\begin{aligned}
    % & L(\psi) = \\
    L(\psi) & = \sum_{i=0}^{K-1} \E_{p_{\theta,t_i}(\rvx_{t_i})q(\rvx_\tau | \hat \rvx_{t_i})} \tfrac{q(\rvx_\tau | \rvx_{t_i})}{q(\rvx_\tau | \rvx_{t_i})} \omega_{\tau}|| f_\psi(\rvx_\tau,\tau) - \hat \rvx_{t_i}||_2^2 \\
    &=  \sum_{i=0}^{K-1} \E_{p_{\theta,t_i}(\rvx_{t_i})q(\rvx_\tau | \rvx_{t_i})} \tfrac{q(\rvx_\tau | \hat  \rvx_{t_i})}{q(\rvx_\tau |\rvx_{t_i})} \omega_{\tau}|| f_\psi(\rvx_\tau,\tau) - \hat \rvx_{t_i}||_2^2. \\
\end{aligned}
\end{equation}
The importance sampling strategy allows us to learn the score in the vicinity of $\rvx_{t_i}$ with lower variance while maintaining an unbiased denoising score matching objective.

\noindent\textbf{Better Teacher Better Student} The proposed objective is data-free, which is a double-edged sword: the advantage lies in not needing to collect image data for training, while the disadvantage is that the pretrained diffusion model limits the performance ceiling. Specifically, the performance upper bound of our method is to perfectly sample from the pre-trained diffusion model. We found that the currently popular SD-v1.5~\cite{rombach2022high}, due to the uneven quality of training data, its generation quality does not align well with human preferences.  To address this issue, we suggest fine-tuning SD-v1.5 on a compact high-quality dataset before distilling student models. This ensures a more capable teacher model, ultimately leading to enhanced student performance. Although recent work~\cite{dmd2} suggest utilizing real data by GANs in distillation, it is less effective and more computationally expensive compared to our proposed pipeline (\cref{sec:main_results}).

\subsection{Enabling Flexible Sampling Steps}
\label{sec:flexible_steps}
Once trained with our proposed objective (Equation \ref{eq:obj}), we can develop a powerful $K$-step generator. While practitioners may need to adjust the number of sampling steps in real-world applications, using the $K$-step generator with fewer steps ($M < K$) yields suboptimal results. This limitation arises because our approach only ensures alignment between intermediate samples in the ODE trajectory and their corresponding diffused distributions.
We observe that existing few-step methods supporting deterministic sampling~\cite{salimans2021progressive,yan2024perflow,sdxllight} suffer from similar limitations, specifically: the distillation target at each step is fixed to samples at predetermined timesteps along the ODE trajectory. This constrains their flexibility in specifying the number of sampling steps and requires multi-stage training to gradually reduce the sampling steps~\cite{sdxllight}.
To address this, we propose that the training objective should be sampling-steps-aware to optimize performance across different step counts. We therefore introduce the following objective:
\begin{equation}
\small
\begin{aligned}
    % \sum_{K \in \mathrm{Steps}} 
    \E_K
    \sum_{i=0}^{K-1}\sum_{\tau = t^{K}_i}^{t^{K}_{i+1}} 
    \lambda_\tau\mathrm{KL}(p_{\theta,\tau|t^{K}_i}(\rvx_\tau|K)  || p_{\phi,\tau}(\rvx_{\tau})),
\end{aligned}
\end{equation}
where $p_{\theta,\tau|t^{K}_i}(\rvx_\tau|K) \triangleq \int q(\rvx_\tau|\rvx_{t^{K}_i}) p{_\theta}(\rvx_{t^{K}_i}|K)d\rvx_{t^{K}_i}$, $K$ is uniformly sampled from a list of desired sampling steps and we let $t_i^{K} = \tfrac{T}{K}i$. Note that the $K$ is injected into both student and fake scores as a condition for guidance.

\noindent\textbf{Why do we need sampling-steps-aware fake score?}  For simplicity, we directly consider the case without noise without loss of generality. Suppose the fake score is shared among $p_{\mathrm{K_1}}(\rvx)$ and $p_{\mathrm{K_2}}(\rvx)$, the optimal fake score becomes:
\begin{equation}
\label{eq:share_fake}
\small
    \frac{p_{\mathrm{K_1}}(\rvx) \nabla_\rvx \log p_{\mathrm{K_1}}(\rvx) + p_{\mathrm{K_2}}(\rvx) \nabla_\rvx \log p_{\mathrm{K_2}}(\rvx)}{p_{\mathrm{K_1}}(\rvx) + p_{\mathrm{K_2}}(\rvx)}
\end{equation}
See \cref{app:derivations} for derivation. It can be seen that even given the optimal fake score, using shared fake scores still introduces bias into training.
This is also the reason that we ensure the diffused intervals do not overlap in the $K$-step learning objective in \cref{eq:obj}. A recent work~\cite{frans2025one} shares a similar idea in sampling-steps aware training, however, their work is limited to point-to-point matching and its extension to text-to-image generation remains unexplored.

\subsection{Surrogate Training Objective}
\label{sec:surrogate_obj}
The Consistency Models (CMs)~\cite{song2023consistency} minimize the following objectives for enforcing self-consistency:
\begin{equation}
\small
   L(\theta) = d(f_\theta(\rvx_{t_n},t_n),\mathrm{sg}(f_\theta(\rvx_{t_{n-1}},t_{n-1})) ),
\end{equation}
The $f_\theta(\rvx_{t_{n-1}},t_{n-1})$ can be regarded as revised samples obtained by denoising from less noisy input. This reveals that both CMs and our proposed method minimize the distance between the generated samples and revised samples while the revised samples are obtained in a different way.

The distance metric is typically chosen as $l_2$~\cite{song2023consistency,luo2023latent}, however, iCT~\cite{song2024improved} has observed that Pseudo-Huber metric is more effective than $l_2$ metric in \cite{song2024improved}. Motivated by their success, and the similarity between CMs and ours in learning generator, we suggest using Pseudo-Huber Metric to form the surrogate training objective as follows:
\begin{equation}
\label{eq:obj_surrogate}
\small
    L(\theta) 
     = \sum_{i=0}^{K-1}\sum_{\tau = t_i}^{t_{i+1}}  \sqrt{||\rvx_{t_i} - \mathrm{sg}(\Tilde{\rvx}_{t_i})||_2^2 +c^2} - c,
\end{equation}
where $c$ is a hyperparameter. Following iCT~\cite{song2024improved}, we let $c = 0.00054\sqrt{d}$ for data with $d$ dimensions.
The corresponding updating gradient for training generator $\theta$ becomes:
\begin{equation}
\small
    \sum_{i=0}^{K-1}\sum_{\tau = t_i}^{t_{i+1}} \frac{[s_\psi(\rvx_\tau,\tau) - s_\phi(\rvx_\tau,\tau)]}{\sqrt{||s_\psi(\rvx_\tau,\tau) - s_\phi(\rvx_\tau,\tau)||_2^2+c^2}-c}  \frac{\partial{\rvx_{t_i}}}{\partial \theta}.
\end{equation}
The Huber metric performs normalization on the gradient, potentially leading to a more stable training process.

\noindent\textbf{Remark} We notice that recent  DMD2~\cite{dmd2} and MMD~\cite{mmd}  also develop few-step generators based on score distillation~\cite{wang2023prolificdreamer,luo2023diff,yin2023one} or moment matching. Our work is more related to DMD2, since MMD applies moment matching for training both generator and fake score in distillation, which does not connect with a certain distribution divergence, and is essentially different from score distillation related to distribution divergence. TDM build on score distillation is more flexible, since we can replace reverse KL by more expensive Fisher divergence~\cite{sid} to achieve better performance (see \cref{tab:fisher} in the Appendix).
Besides, DMD2 always aims to predict clean samples at each step which ignores the intermediate stage of diffusion ODE trajectory, while MMD uses DDPM sampler~\cite{ho2020denoising} from noisy real data for prediction in training which introduces large stochasticity. In contrast, our objective is aimed at performing \textit{deterministic} trajectory distillation at the distribution level. Their designs increase the learning difficulty of both the generator and fake score, and are unnecessarily challenging and sub-optimal for few-step sampling. Moreover, their fake score are fully shared among different steps, this is flawed in theory and further increases the learning difficulty. As a result, DMD2 has to train fake scores multiple times ($\geq 5$) at each iteration for better performance.
Lastly, our work builds non-trivial connection between trajectory distillation and distribution matching, delivering a new unified distillation paradigm.

\begin{table*}[!t]
        \centering
        \caption{Comparison of machine metrics on text-to-to-image generation across state-of-the-art methods. \shortname-unify-SFT is initialized from fine-tuned SD-v1.5 and \shortname-unify-GAN is initialized from original SD-v1.5.
        HFL denotes human feedback learning which might hack the machine metrics. We \textbf{highlight} the best among distillation methods.}
        \label{tab:main_t2i}
        \vspace{-2mm}
        \resizebox{.85\linewidth}{!}{
        \begin{tabular}{lccccccccccc}
        \toprule
             \multirow{2}{*}{Model} & \multirow{2}{*}{Backbone} & \multirow{2}{*}{HFL} & \multirow{2}{*}{Steps} & \multicolumn{5}{|c|}{HPS$\uparrow$} & \multirow{2}{*}{Aes$\uparrow$}  & \multirow{2}{*}{CS$\uparrow$}   & \multirow{2}{*}{\textbf{Image-Free?}} \\
              &  &  &  & \multicolumn{1}{|c}{Animation} & Concept-Art & Painting & Photo & \multicolumn{1}{c|}{Average} &  & \\
        \midrule
             \rowcolor{gray!20}
             Base Model (CFG = 3.5) & SD-v1.5 & No & 25 & 26.29 & 24.85 & 24.87 & 26.01 & 25.50 & 5.49 & 33.03  &  \\
             \rowcolor{gray!20}
             Base Model + Fine-tuning (CFG = 3.5) & SD-v1.5 & No & 25 & 31.10 & 29.88 & 29.53 & 28.94 & 29.86 & 5.85 & 33.68 &  \\
             InstaFlow~\citep{liu2023insta} &  SD-v1.5  & No & 1  & 23.17 & 23.04 &22.73 &22.97 & 22.98 & 5.27  & 30.04 & \ding{55}  \\
             PeRFlow~\citep{yan2024perflow} &  SD-v1.5  & No & 1 & 12.37 & 13.50 & 13.64 & 11.53 & 12.76 & 4.47 & 15.49 & \ding{55} \\
             PeRFlow~\citep{yan2024perflow} &  SD-v1.5  & No & 2 & 19.75 & 19.43 & 19.41 & 18.40 & 19.25 & 4.91 & 25.83 & \ding{55} \\
             Hyper-SD~\citep{ren2024hypersd} &  SD-v1.5  & Yes & 1  & 28.65 & 28.16 & 28.41 & 26.90 & 28.01 & 5.64 & 30.87  & \ding{55} \\
             \textbf{\shortname-unify-GAN (Ours)} &  SD-v1.5 & No & 1  & 29.80 & 28.66 & 28.82 & 26.80  & 28.54 & 5.97  & 31.89 & \ding{55} \\
             \textbf{\shortname-unify-SFT (Ours)} &  SD-v1.5 & No & 1    & \textbf{29.85} & \textbf{28.90} & \textbf{29.22} & \textbf{27.62}  & \textbf{28.90} & \textbf{6.02}  & \textbf{32.12} & \ding{55} \\
             \midrule
             LCM-dreamshaper~\citep{luo2023lcmlora} &  SD-v1.5 & No & 4    & 26.51 & 26.40 & 25.96 & 24.32 & 25.80 & 5.94 & 31.55 & \ding{55} \\
             PeRFlow~\citep{yan2024perflow} &  SD-v1.5  & No & 4  & 22.79 & 22.17 & 21.28 & 23.50 & 22.43 & 5.35 & 30.77 & \ding{55} \\
             TCD~\citep{tcd} &  SD-v1.5 & No & 4    & 23.14 & 21.11 & 21.08 & 23.62 & 22.24 & 5.43 & 29.07 & \ding{55} \\
             Hyper-SD~\citep{ren2024hypersd} &  SD-v1.5  & Yes & 4  & 31.06 & 30.01 & 30.47 & 28.97 & 30.24 & 5.78  & 31.49 & \ding{55} \\
             DMD2~\cite{dmd2} & SD-v1.5 & No & 4 & 30.69 & 29.43 & 29.75 & 28.07 & 29.49 & 5.91 & 31.53 & \ding{55} \\
             \textbf{\shortname-unify-GAN (Ours)} &  SD-v1.5 & No & 4 & 32.04 & 30.86 & 31.06 & 29.35 & 30.83 & 6.07 & 32.40 & \ding{55}\\
             \textbf{\shortname-unify-SFT (Ours)} &  SD-v1.5 & No & 4    & \textbf{32.40} & \textbf{31.65} & \textbf{31.35} & \textbf{29.86}  & \textbf{31.31} & \textbf{6.08} & \textbf{32.77} & \ding{55} \\
             \midrule
             \midrule
             \rowcolor{gray!20}
             Base Model-1024 (CFG=7.5) & SDXL & No & 25 & 34.66 & 33.70	 & 33.43 & 30.95 & 33.19	 & 6.17 & 36.28 & \\
             TCD~\cite{tcd} & SDXL & No & 4 & 29.65 & 27.50 & 27.98 & 26.13 & 27.81 & 5.88  & 33.42 & \ding{55}  \\
             LCM~\cite{luo2023lcmlora} & SDXL & No & 4 & 30.79 & 29.38 & 29.60 & 27.87& 29.41 & 5.84 & 34.84 & \ding{55} \\
             SDXL-Turbo-512~\cite{sauer2023adversarial} & SDXL & No & 4 &  32.54 & 31.03  & 31.04 & 28.60 & 30.80 & 5.81  & 35.03  & \ding{55} \\       
             SDXL-Lighting~\cite{sdxllight} & SDXL & No & 4 & 34.20 & 32.97 & 33.15 & 30.52 & 32.71 & 6.23 & 34.62  & \ding{55} \\  
             Hyper-SD~\cite{ren2024hypersd} & SDXL & Yes & 4 &  35.58 &  34.54 & 34.54 & 31.90 & 34.14 & 6.18 & 34.27  & \ding{55} \\   
             DMD2~\cite{dmd2} & SDXL & No & 4 &  32.87 &  31.56 & 31.01 & 30.39 & 31.46 & 5.88 & 35.51 & \ding{55} \\   
             \textbf{\shortname (Ours)} & SDXL & No & 4 &  \textbf{36.42} &  \textbf{35.34}	 &  \textbf{35.51} & \textbf{32.25}  & \textbf{34.88}	 & \textbf{6.28} & \textbf{36.08} & \checkmark\\   
             \midrule
             \midrule
            \rowcolor{gray!20}
            Base Model-1024 (CFG=3.5) & PixArt-$\alpha$ & No & 25 & 33.54 & 32.35 & 32.00 & 30.93 & 32.21 & 6.23 & 34.11 & \\
            YOSO-512~\cite{yoso} &  PixArt-$\alpha$  & No & 4   & 31.40 & 31.18 & 31.26 & 28.15 & 30.60 & 6.23 & 31.83 & \ding{55} \\
            LCM-1024~\citep{luo2023latent} &  PixArt-$\alpha$  & No & 4   & 31.96 & 30.60 & 30.70 & 28.92 & 30.55 & 6.17  & 33.49  & \ding{55} \\
            \textbf{\shortname-1024 (Ours)} &  PixArt-$\alpha$  & No & 4  & \textbf{34.61} & \textbf{33.54} & \textbf{33.45} & \textbf{31.23} & \textbf{33.21} & \textbf{6.42} & \textbf{33.66} & \checkmark \\
        \bottomrule
        \end{tabular}
                }
        \vspace{-2mm}
\end{table*}

\begin{table*}[!ht]
        \centering
        \caption{Comparison of machine metrics on integrating LoRA into \textit{unseen customized models} across state-of-the-art methods. HFL denotes human feedback learning which might hack the machine metrics. The FID is computed between teacher samples and student samples for measuring the \textbf{\textit{style preservation}}. We \textbf{highlight} the best among distillation methods. }
        \vspace{-2mm}
        \label{tab:main_lora}
        \resizebox{0.85\linewidth}{!}{
        \begin{tabular}{lccccccccccc}
        \toprule
             \multirow{2}{*}{Model} & \multirow{2}{*}{Backbone} & \multirow{2}{*}{HFL} & \multirow{2}{*}{Steps} & \multicolumn{5}{|c|}{HPS$\uparrow$} & \multirow{2}{*}{Aes$\uparrow$}  & \multirow{2}{*}{CS$\uparrow$} & \multirow{2}{*}{FID$\downarrow$} \\
              &  &  &  & \multicolumn{1}{|c}{Animation} & Concept-Art & Painting & Photo & \multicolumn{1}{c|}{Average} &  & & \\
            \midrule
             \rowcolor{gray!20}
             Realistic & SD-v1.5 & No & 25 & 31.28 & 30.08 & 29.73 & 29.02 & 30.03 & 5.88  & 34.41 & - \\
             LCM~\citep{luo2023lcmlora} &  SD-v1.5 & No & 4 & 28.85 & 27.05 &28.08  & 26.91 & 27.72 & 5.79 & 30.95 & 26.89 \\
             PeRFlow~\citep{yan2024perflow} &  SD-v1.5  & No & 4 & 26.66 & 25.73 &25.83  & 24.54 & 25.69 & 5.59 & 31.93 & 25.84\\
             TCD~\citep{tcd} &  SD-v1.5  & No & 4  & 28.42 & 26.21 & 26.85 & 26.33 & 26.95 & 5.82  & 31.28 & 28.65\\
             Hyper-SD~\citep{ren2024hypersd} &  SD-v1.5  & Yes & 4  & 31.31 & 30.35 & 30.90 & 28.86 & 30.36 & 5.97  & 32.19 & 37.83\\
             \textbf{\shortname (Ours)} &  SD-v1.5 & No & 4  & \textbf{32.33} & \textbf{31.29} &\textbf{31.49} &\textbf{29.78} & \textbf{31.22} & \textbf{6.04}  & \textbf{32.63} & \textbf{20.23} \\
             \midrule
             \rowcolor{gray!20}
             Dreamshaper & SD-v1.5 & No & 25 & 31.90 & 30.19 &30.26  & 29.28 &  30.41 & 6.02 & 34.20 & - \\
             LCM~\citep{luo2023lcmlora} &  SD-v1.5 & No & 4 & 29.78 & 28.25 & 29.11 & 27.23 & 28.59 & 5.98  & 31.10 & 25.36 \\
             PeRFlow~\citep{yan2024perflow} &  SD-v1.5  & No & 4 & 27.37 & 26.50 &26.66 &25.16  & 26.42 & 5.74  & 32.15 & 23.49\\
             TCD~\citep{tcd} &  SD-v1.5  & No & 4 & 29.46 & 27.49 & 28.26 & 26.42 & 27.91 & 6.01  & 31.28 & 28.65\\
             Hyper-SD~\citep{ren2024hypersd} &  SD-v1.5  & Yes & 4  & 32.05 & 30.98 & 31.37 & 28.87 & 30.82 & 6.13 & 31.54 & 38.70 \\
             \textbf{\shortname (Ours)} &  SD-v1.5 & No & 4   & \textbf{32.91} & \textbf{31.73} & \textbf{32.18} & \textbf{29.95}& \textbf{31.37} & \textbf{6.22} & \textbf{32.30} & \textbf{20.44}\\
       \bottomrule
        \end{tabular}
                }
        \vspace{-2mm}
\end{table*}

\vspace{-2mm}
\section{Experiment}
\vspace{-2mm}

\spara{Experiment Setting}
The distillation is conducted on the JourneyDB dataset~\citep{pan2023journeydb} solely with its prompts, since our method is image-free. 
Using the technique proposed in \cref{sec:flexible_steps}, we develop \shortname-unify on SD-v1.5 that enables flexible sampling steps. We compared two variants of utilizing real data for \shortname-unify: 1) fine-tuning SD-v1.5 before distillation, termed \shortname-unify-SFT; 2) applying GANs during distillation, termed \shortname-unify-GAN.
We use LAION-AeS-6.5+~\cite{schuhmann2022laion} as the compact high-quality dataset. 
More experiment details can be found in the \cref{app:exp_details}.

\spara{Evaluation} We employ Aesthetic Score (AeS)~\citep{schuhmann2022laion} to evaluate image quality and adopt the Human Preference Score (HPS) v2.1~\citep{wu2023human} to evaluate the image-text alignment and human preference. 
Additionally, we include CLIP score (CS)~\citep{hessel2021clipscore} to provide a more comprehensive evaluation. We mainly compare our model against the \texttt{open-source} state-of-the-art (SOTA) models.

\begin{figure*}[!t]
    \centering
    \includegraphics[width=1\linewidth]{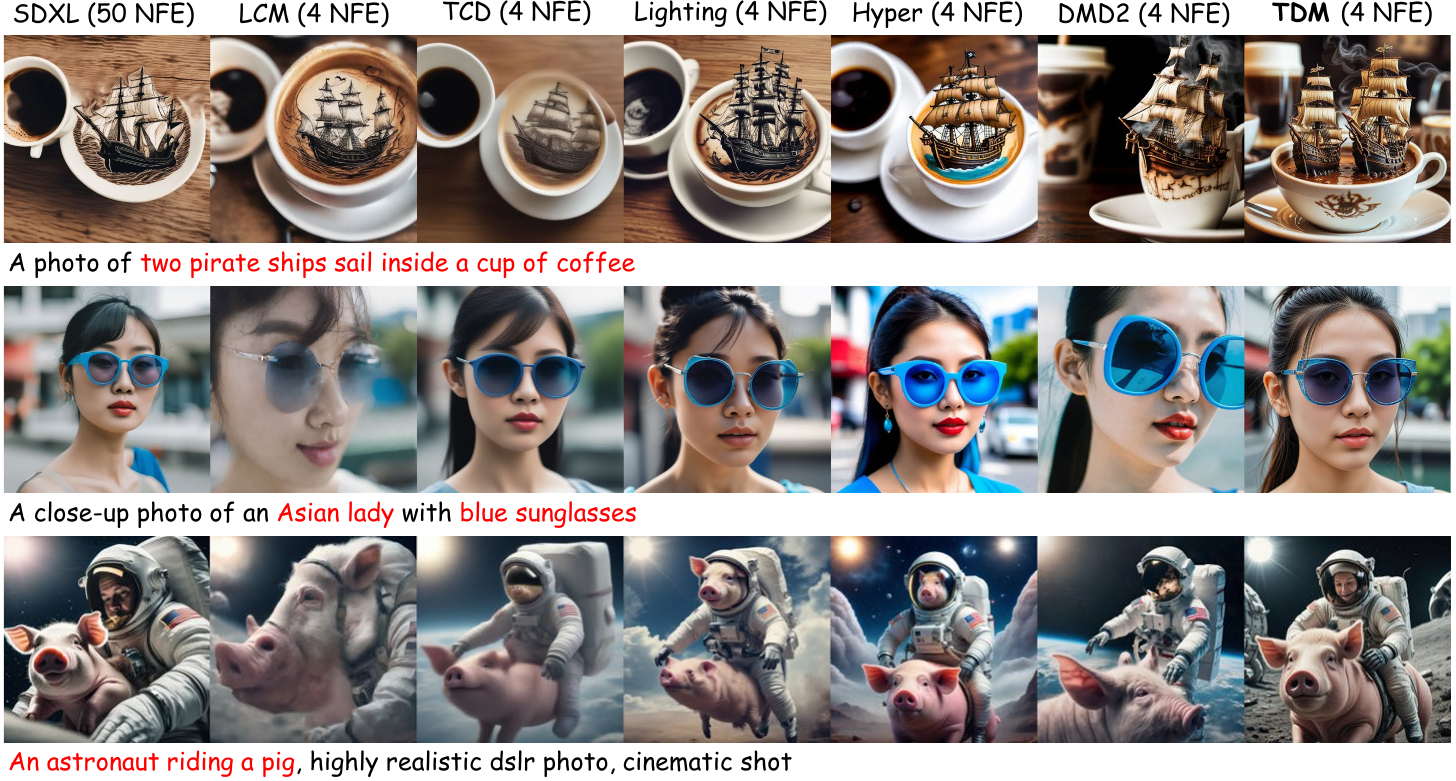}
    \vspace{-6mm}
    \caption{Qualitative comparisons of \shortname against most competing methods on SDXL. All images are generated by the same initial noise.}
    \label{fig:visual_compare}
    \vspace{-2mm}
\end{figure*}
\begin{figure*}[!ht]
    \centering
    \includegraphics[width=1\linewidth]{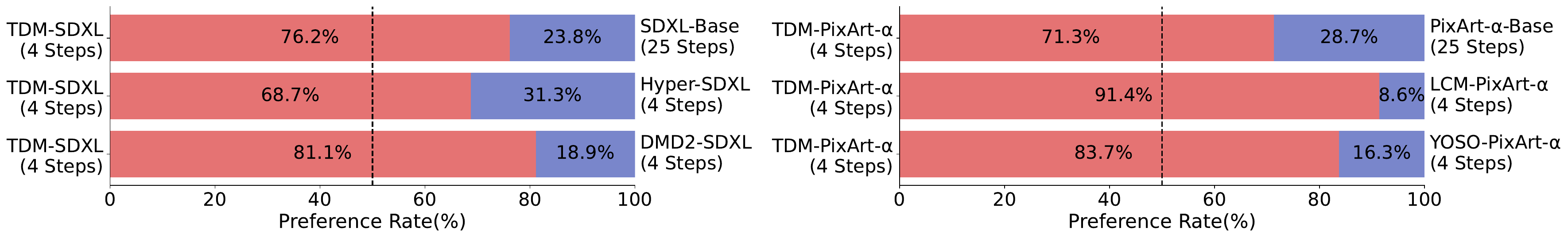}
    \vspace{-7.5mm}
    \caption{The user study about the comparison between our method and the most competing methods.}
    \label{fig:user_prefer}
    \vspace{-4mm}
\end{figure*}

\vspace{-1mm}
\subsection{Main Results}
\label{sec:main_results}
\vspace{-1mm}

\spara{Quantitative Results}  
We evaluate our method against existing baselines, including fine-tuned models and prior distillation approaches, using the HPS benchmark. As demonstrated in \cref{tab:main_t2i}, our approach consistently achieves superior performance across all architectural variants and evaluation metrics. While Hyper-SD shows competitive results, its performance relies on human feedback learning, which may artificially inflate machine metrics. In contrast, our method achieves better results without such techniques.
Notably, our unified model provides SOTA performance on various sampling steps by deterministic sampling, highlighting the effectiveness of our sampling-steps aware objective. For comparison, PeRflow, the previous SOTA in supporting deterministic sampling, performs significantly worse in 1 or 2-step generation. 

\spara{Data Introduction Strategy} We Compare two approaches for incorporating high-quality data: (1) fine-tuning DMs before TDM distillation, and (2) using GANs during distillation. The \shortname-unify-SFT clearly (31.31 HPS) outperforms \shortname-unify-GAN in terms of 4-step generation, while being more stable and computationally efficient. We note that \shortname-unify-GAN requires an extra 30\%  of the overall computational cost compared to \shortname-unify-SFT's 3 A800 days (includes fine-tuning and distillation), or its 4-step performance degrades to 29.80 HPS. This indicates the effectiveness of our proposed strategy in utilizing real data.

\spara{Integrating LoRA into Customized Models} 
We examine the efficacy of LoRA integration across unseen base models, evaluating both generation quality and style preservation. Style preservation is assessed by the FID~\cite{heusel2017gans} between the teacher samples and student samples. We train TDM-LoRA for original SD-v1.5 in an image-free way. As shown in \cref{tab:main_lora}, our method significantly outperforms existing approaches. While Hyper-SD achieves competitive HPS and AeS scores, its high FID scores indicate poor style preservation relative to the original base models. Visual comparisons in \cref{app:addtional_visual} further demonstrate our method's superior capability in maintaining both high-fidelity generation and faithful style preservation.

\spara{Qualitative Comparison} We present the qualitative comparison in \cref{fig:visual_compare}. It is clear that our method has better visual quality and text-image alignment compared to competing baselines and even the teacher diffusion with 25 steps.

\begin{table}[!t]
    \centering
    \caption{Comparison on training cost across backbones and methods. 
    \shortname-unify-SFT's cost includes the fine-tuning stage. 
    \shortname-unify is more costly as it requires training various sampling steps.
    }
    \vspace{-2.5mm}
    \label{tab:cost}
    \resizebox{0.9\linewidth}{!}{
    \begin{tabular}{lcccc}
    \toprule
     Method & Backbone & NFE$\downarrow$ & HPS$\uparrow$ & Training Cost \\ 
     \midrule
     DMD2~\cite{dmd2} & SD-v1.5 & 4 &  31.53 & 30+ A800 Days \\ 
     \textbf{\shortname-unify-GAN} & SD-v1.5 & 4 & 32.40 & 4 A800 Days \\ 
     \textbf{\shortname-unify-SFT} & SD-v1.5 &4 & \textbf{32.77} & \textbf{3 A800 Days} \\ 
     \midrule
     \midrule
     LCM~\cite{luo2023lcmlora} & SDXL & 4 & 29.41 & 32 A100 Days \\  
     DMD2~\cite{dmd2} & SDXL & 4 & 31.46 & 160 A100 Days \\ 
     \textbf{\shortname} & SDXL & 4 & \textbf{34.88} & \textbf{2 A800 Days} \\
     \midrule
     \midrule
     LCM~\cite{luo2023lcmlora} & PixArt-$\alpha$ & 4 & 30.55  & 14.5 A100 Days\\
     YOSO~\cite{yoso} & PixArt-$\alpha$ & 4 & 30.60  & 10 A800 Days \\
     \textbf{\shortname} & PixArt-$\alpha$ & 4 & \textbf{32.01} &   \textbf{2 A800 Hours} \\
    \bottomrule
    \end{tabular}
    }
    \vspace{-3mm}
\end{table}

\begin{figure}[t]
    \centering
    \includegraphics[width=1\linewidth]{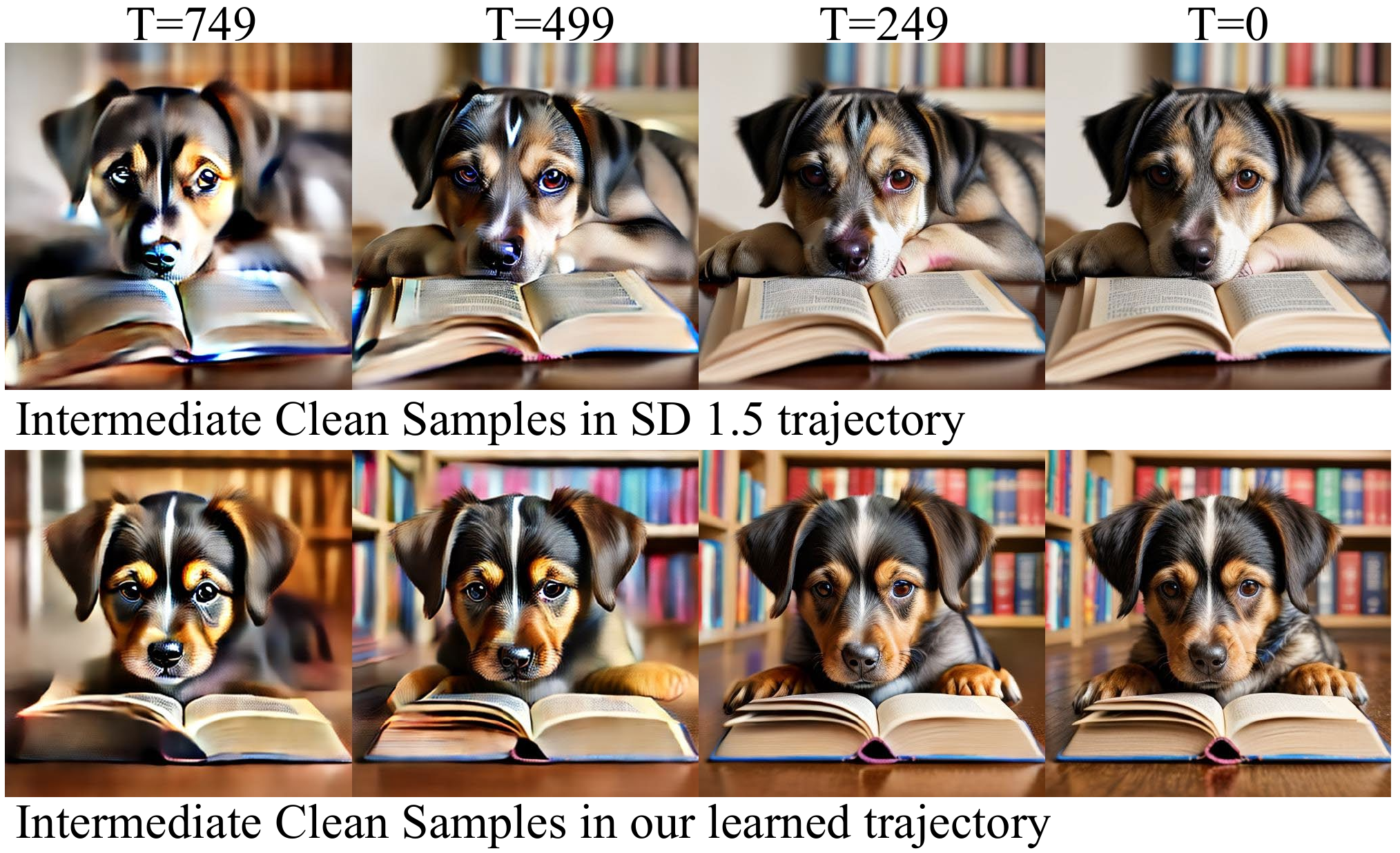}
    \vspace{-6mm}
    \caption{The visualization of ODE trajectory with clean samples at different timesteps. It is clear that our method suffers less from the CFG artifact and has better visual quality. The prompt is ``A dog reading a book". See \cref{app:add_imgs} for more visualizations.}
    \label{fig:traj_compare}
    \vspace{-5mm}
\end{figure}

\spara{User Study} To further verify the effectiveness of our proposed method, we conduct an extensive user study across different backbones. The results in \cref{fig:user_prefer} shows that our method clearly outperforms the teacher diffusion and other most competing methods.

\spara{Surpassing Teacher Diffusion Without Requiring Extra Data} We observe a notable phenomenon: our proposed method outperforms both the original multi-step diffusion models and their fine-tuned variants across machine metrics and user study, without requiring extra data. The reason behind this phenomenon lies in our distillation target: the sequence of pre-trained diffused distributions ${p_{\phi,t}(\rvx_t)}$. While multi-step DMs generate samples by solving PF-ODE, they inevitably introduce numerical errors, making it challenging to sample perfectly from the distribution sequence. In contrast, our distillation objective does not require sampling from the teacher distribution, resulting in superior distilled performance despite using 10x fewer NFEs.

\spara{\shortname Learned ODE Trajectory at Distribution Level Well} We present the middle clean samples in the trajectory of Teacher DMs and our \shortname in \cref{fig:traj_compare}. It can be seen that \shortname learns the trajectory well. Notably, we found that \shortname suffers less from the CFG artifact and has better details at each step, finally forming better visual quality.

\spara{Training Cost} We highlight that it is extremely efficient to train a 4-step generator that surpasses the teacher model using our proposed \shortname. This is due to: 1) We support deterministic sampling, which inherently possesses good few-step capabilities; 
2) Our distillation target is to simulate the teacher’s trajectory at the distribution level, each step only requires partial denoising and does not always need to predict a clean image, which reduces the difficulty of learning.
As shown in \cref{tab:cost}, our method substantially outperforms existing approaches while drastically reducing computational requirements. Notably, on SDXL and PixArt-$\alpha$, our method requires only 1.25\% and 0.57\% of the training costs of DMD2 and YOSO, respectively, demonstrating exceptional training efficiency.

\begin{figure}[!t]
    \centering
    \includegraphics[width=1\linewidth]{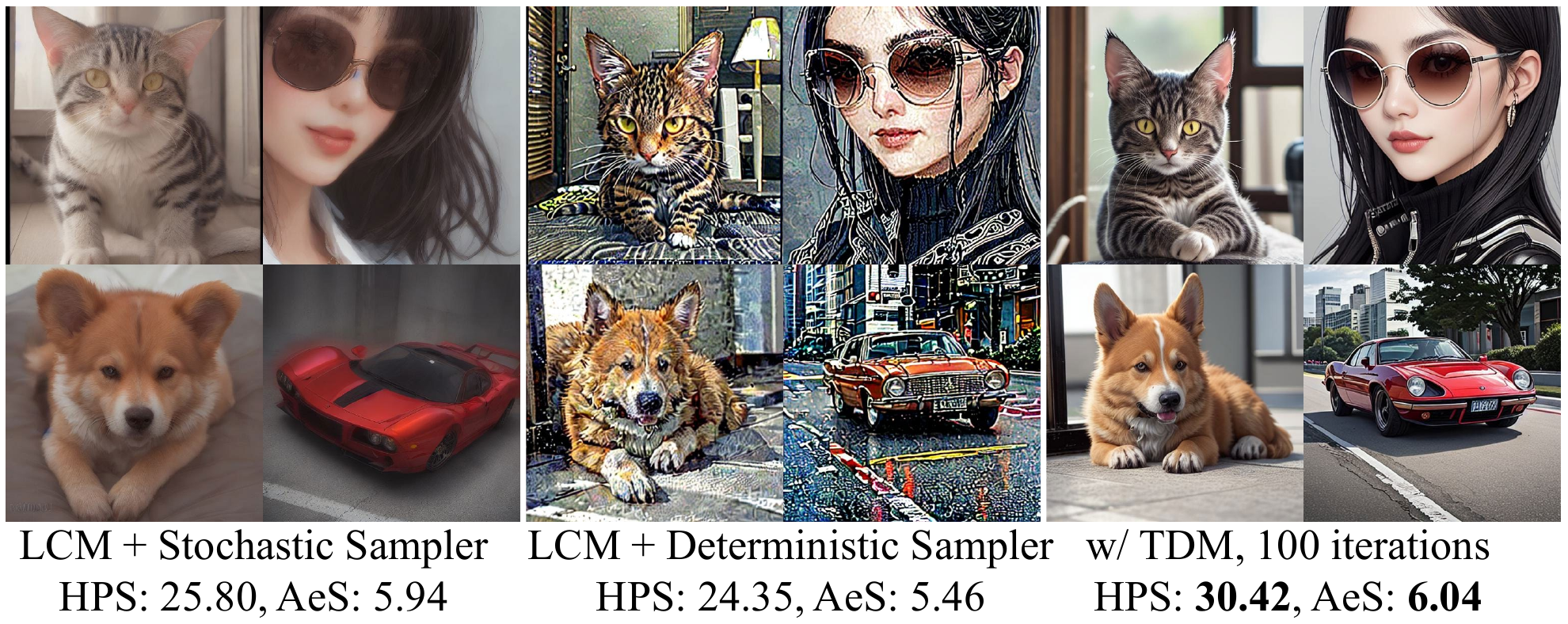}
    \vspace{-6.5mm}
    \caption{4 step generation from LCM and our method initialized by LCM. Our method can recover LCM from poor deterministic sampling via merely 100 training iterations.}
    \label{fig:recover}
    \vspace{-4mm}
\end{figure}

\spara{Fixing LCMs from Poor Deterministic Sampling} 
While deterministic sampling is crucial for our method's efficiency and performance, popular models like LCM~\cite{luo2023latent} are limited to stochastic sampling, potentially compromising their effectiveness. We investigated whether our approach could recover deterministic sampling capability in these models. Using LCM-dreamshaper as our baseline, we demonstrate that our method can successfully recover deterministic sampling functionality within just 100 training iterations, leading to superior 4-step performance (\cref{fig:recover}). 

\subsection{Additional Results on Text-to-Video Acceration}
\begin{table}[!t]
    \centering
    \caption{Evaluation of text-to-video on Vbench.}
    \vspace{-2mm}
    \label{tab:t2v}
    \resizebox{.9\linewidth}{!}{
    \begin{tabular}{lcccc}
    \toprule
     Method & Total Score$\uparrow$ & Quality Score$\uparrow$ & Semantic Score$\uparrow$ \\ 
     \midrule
     CogVideoX-2B & 80.91 & 82.18 & 75.83 \\
     TDM (4 NFE) & \textbf{81.65} & \textbf{82.66} & \textbf{77.64} \\
    \bottomrule
    \end{tabular}
    }
    \vspace{-3mm}
\end{table}
Our proposed \shortname can be extended to text-to-video generation, benefiting from its quick convergence and data-free property. We distill our 4-step TDM from CogVideoX-2B~\cite{yang2024cogvideox}. We evaluate our method on the widely used VBench~\cite{huang2023vbench}. Results shown in \cref{tab:t2v} show that TDM surpasses the teacher CogVideoX-2B by a notable margin while using just 4 NFE.

\subsection{Ablation Study}

\begin{table}[!t]
    \centering
    \caption{Comparison on HPS across variants based on SD-v1.5.}
    \vspace{-3mm}
    \label{tab:ablation}
    \resizebox{.9\linewidth}{!}{
    \begin{tabular}{lccc}
    \toprule
     Method & 1 Step & 2steps & 4 Steps \\ 
     \midrule
     \shortname-unify & \textbf{28.90} & \textbf{30.52} & 31.31\\
     \midrule
     w/o Conditioned on Sampling Steps & 26.11 & 29.15 & 29.39\\
     w/o Surrogate Training Objective & 28.23 & 30.20 & 30.85\\ % better checkpoint
     w/o unify training (TDM-4step) & 20.81 & 29.08 & \textbf{31.35}\\
     \midrule
     \midrule
     \shortname-4Step & / & / & \textbf{31.35}\\
     \midrule
     w/o Importance Sampling & / & / & 29.27\\
    \bottomrule
    \end{tabular}
    }
    % \vspace{-6mm}
\end{table}

We conduct comprehensive ablation studies on our proposed components, with results shown in \cref{tab:ablation}. See \cref{app:add_ablation} for Setting details and additional ablation studies.
\spara{Unify Training} Without the unify training, performance drops notably in fewer steps (20.81 vs. 28.90 HPS for 1-step, 29.08 vs. 30.52 HPS for 2-steps).

\spara{Importance Sampling} Adopting importance sampling for fake score learning significantly improves the 4-step generator's performance (31.35 vs. 29.27 HPS). Without it, we observe degraded mode coverage and training stability.

\spara{Sampling-Steps-Aware Training} Conditioning on sampling steps enables flexible step selection during inference. Without this conditioning, performance drops notably (26.11 vs. 28.90 for 1-step, 29.39 vs. 31.31 for 4-steps), likely due to distribution interference across different steps by sharing fake score as we analyzed in \cref{sec:flexible_steps}.

\spara{Surrogate Training Objective} Our surrogate objective, inspired by Consistency Models, enhances the generator's performance (31.31 vs. 30.85 HPS), validating the effectiveness of this design choice.

\begin{figure}[t]
    \centering
    \includegraphics[width=1\linewidth]{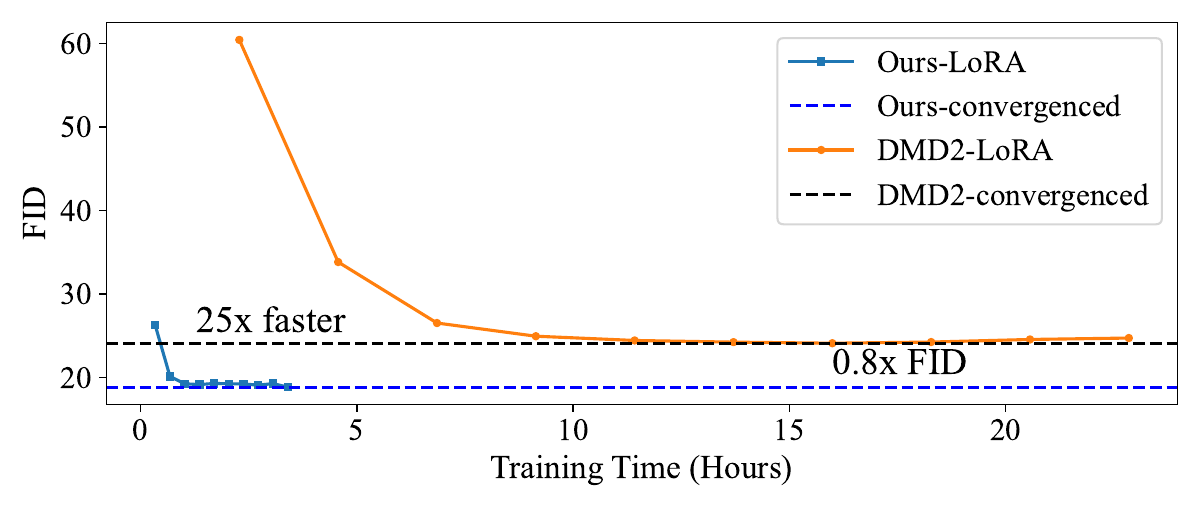}
    \vspace{-7mm}
    \caption{Comparison to DMD2 under LoRA fine-tuning.}
    \label{fig:compare_curve}
    \vspace{-5mm}
\end{figure}

\spara{Comparison to DMD2} It is important to distinguish our method from DMD2~\cite{dmd2}. 
While DMD2 employs score distillation~\cite{wang2023prolificdreamer} with direct image prediction at multiple timesteps, our approach focuses on distribution-level trajectory alignment which enables easier learning and faster convergence. 
To gain more intuitive insight into the superiority of our proposed \shortname, we conducted experiments in distilling original SD v1.5 via LoRA.  We present the FID between teacher samples and student samples at the training process in \cref{fig:compare_curve}. We observed that the optimal FID of our method is 20\% lower than that of DMD2, and it only requires 4\% of the training time to reach the FID convergence of DMD2, which demonstrates the significant advantages of our \shortname in terms of training efficiency and performance.

\vspace{-3.5mm}
\section{Conclusion}
\vspace{-1mm}
We propose \shortname, a new distillation paradigm that unifies trajectory distillation and distribution matching. Our method can surpass the teacher in a data-free manner, achieving new SOTA performance with fewer steps. This is achieved with extremely low training costs. Specifically, we only need 2 A800 hours to distill a 4-step generator with \shortname that outperforms its teacher PixArt-$\alpha$ in both quantitative and human evaluations. 
This shows the efficiency and effectiveness of our method, paving the way for easier and more inclusive future research on diffusion distillation.

{
    \small
    \bibliographystyle{ieeenat_fullname}
    \bibliography{main}
}

\clearpage
\setcounter{page}{1}
\maketitlesupplementary
\appendix

\section{Training Algorithm}
We present the algorithm for distilling K-step \shortname in \cref{algo:training_tdm}.

\begin{algorithm}[h]
\caption{Trajectory Distribution Matching.}
\label{algo:training_tdm}
\begin{algorithmic}[1]

\Require learning rate $\eta$, desired sampling steps $K$, total iterations $N$, real score $f_\phi$.
\Ensure optimized models $f_\theta$, $f_\psi$. 
\State Initialize weights $\{ \theta, \psi \}$ by $\phi$; 

\For{$i \leftarrow 1$ {\bfseries to} $N$}

    \State Sample noise $\epsilon$ from standard normal distribution;
    \State Sample $\{\rvx_{t_i}\}_{i=0}^{K-1}$ with initialized noise $\epsilon$ from generator $f_\theta$ by $K$ steps via ODESolver. 
    \State Sample $\rvx_{t_m}$ from $\{\rvx_{t_i}\}_{i=0}^{K-1}$.
    \State Sample Timesteps $t$ from $t_m$ to $t_{m+1}$.
    \State Obtain noisy samples $\rvx_t \sim q(\rvx_t|\rvx_{t_m})$

    % \State
    \State \texttt{\textcolor{purple}{\# update fake score}}
    \State Compute Loss $\mathcal{L}_\psi$ following  \texttt{Line 234}.
    \State $\psi \leftarrow \psi - \eta\nabla_{\psi}\mathcal{L}_\psi$;

    % \State
    \State \texttt{\textcolor{purple}{\# update Generator}}
    \State Compute Loss $\mathcal{L}_\theta$ following \cref{eq:obj_surrogate}.
    \State $\theta \leftarrow \theta - \eta\nabla_{\theta}\mathcal{L}_\theta$;
\EndFor
\end{algorithmic}
\end{algorithm}

\section{Derivations}
\label{app:derivations}

\paragraph{Derivations of \cref{eq:share_fake}.} 

Given the case without noise, the learning objective of the fake score $s_\psi$ by score matching is:
\begin{equation}
\begin{aligned}
    L(\psi) = & \E_{p_{K_1}(\rvx)}||\nabla \log p_{K_1}(\rvx) - s_\psi(\rvx) ||_2^2 \\
    & + \E_{p_{K_2}(\rvx)}||\nabla \log p_{K_2}(\rvx) - s_\psi(\rvx) ||_2^2.
\end{aligned}
\end{equation}
Its gradient can be computed as follows:
\begin{equation}
\begin{aligned}
    \nabla_\psi L(\psi) = 
    \int [2 p_{K_1}(\rvx) (\nabla \log p_{K_1}(\rvx) - s_\psi(\rvx)) \\ + 2 p_{K_2}(\rvx) (\nabla \log p_{K_2}(\rvx) - s_\psi(\rvx))]\frac{\partial s_\psi(\rvx) }{\partial \psi} d\rvx
\end{aligned}
\end{equation}
The global minimum is achieved when $\nabla_\psi L(\psi) = \mathbf{0}$. It is clear that when
\begin{equation}
\label{eq:optimal_share_fake}
\small
    s_\psi(\rvx) = \frac{p_{\mathrm{K_1}}(\rvx) \nabla_\rvx \log p_{\mathrm{K_1}}(\rvx) + p_{\mathrm{K_2}}(\rvx) \nabla_\rvx \log p_{\mathrm{K_2}}(\rvx)}{p_{\mathrm{K_1}}(\rvx) + p_{\mathrm{K_2}}(\rvx)},
\end{equation}
we have $\nabla_\psi L(\psi) = \mathbf{0}$. Hence the optimal fake score is given in \cref{eq:optimal_share_fake}.

% \vspace{-2mm}
\section{Additional Related Works}
\label{sec:related_work}
% \vspace{-2mm}
Recently, there mainly have been two lines in diffusion distillation: Trajectory Distillation and Distribution Matching. Trajectory distillation tries to distill a few-step student model by simulating the generative process of DMs.  
These methods typically predict the multi-step solution of PF-ODE solver by one step~\cite{on_distill,salimans2021progressive,liu2023insta,yan2024perflow}. 
Consistency family~\cite{song2023consistency,song2024improved,luo2023latent,ctm,tcd} enforces self-consistency. 
These methods suffer from numerical errors when solving pre-trained PF-ODE.
Distribution Matching tries to distill student models via match at the distribution level. Diffusion-GAN hybrid models~\cite{ufogen,sauer2023adversarial,kang2024diffusion2gan} have been proposed for this aim, however, stabilizing GAN training requires real data, careful architecture design, and auxiliary regression losses.
Another promising way for distribution matching is through score distillation~\cite{wang2023prolificdreamer,luo2023diff,sid,yin2023one,nguyen2024swiftbrush}.
These methods typically ignore the intermediate steps of the trajectory.
Our method explores trajectory distillation at the distribution level, enjoying the best of two worlds. Although Hyper-SD~\cite{ren2024hypersd} explores combining trajectory distillation and distribution matching, their works treat these two techniques as distinguished parts, requiring multiple training objectives and multiple training stages. In contrast, our proposed objective naturally unifies trajectory distillation and distribution matching, providing a highly efficient and effective distillation method. 
Besides, motivated by the similarity between consistency models and our proposed method in learning generator, we propose a surrogate training objective, introducing the Huber metric into training. 
This leads to better performance and potentially encourages the community to explore other types of distance metrics in distilling via distribution matching. Recently, DMD2~\cite{dmd2} also explored distilling a few-step generator via distribution matching. However, our method is fundamentally different from their work. Their work tries to predict clean images at different timesteps, ignoring alignment with the teacher's trajectory. This leads to harder learning and slower convergence. In particular, in distilling 4-step SDXL, we only require 1.25\% of the training cost for DMD2, while achieving significantly better performance. Besides, a recent work MMD~\cite{mmd} also developed a multi-step generator based on a similar style with score distillation. However, our work is essentially different from them, since MMD applies moment matching for diffusion distillation, while we propose a new distillation paradigm that unifies trajectory distillation and distribution matching. Specifically, MMD employs moment matching to train both generator and fake ``score" which is fundamentally different from score matching as noted in their paper~\cite{mmd}; In contrast, we use reverse KL to train the generator and score matching to train the fake score.  Moreover, MMD uses ancestral sampling (DDPM sampler~\cite{ho2020denoising}) from noisy real data in training, which introduces large stochasticity in intermediate samples, is sub-optimal for few-step sampling. In contrast, we use deterministic sampling from noise which is image-free, more effective for fewer-step sampling, and builds a non-trivial connection with trajectory distillation.

Additionally, a concurrent work~\cite{frans2025one} explored flexible deterministic sampling but was limited by point-to-point self-consistency and required training the generator at arbitrary timesteps, challenging model capacity. Its extension to text-to-image generation also remains unclear. In contrast, our method trains the generator at only K timesteps with distribution-level matching, enabling more effective flexible deterministic sampling. We further achieve state-of-the-art performance in text-to-image generation.

\section{Experiment details}
\label{app:exp_details}
We use the AdamW optimizer for both the generator and fake score. By default, the $\beta_1$ is set to be 0, the $\beta_2$ is set to be 0.999.

\paragraph{SD-v1.5} We adopt a constant learning rate of 2e-6 for the generator and 2e-5 for the fake score. We apply gradient norm clipping with a value of 1.0 for both the generator and fake score. We use batch size 256. We set the CFG as 3.5. Generally, the training is done within 20k iterations for unified training and within 3k iterations for specific 4-step training.

\paragraph{SDXL} We adopt a constant learning rate of 1e-6 for the generator and 5e-6 for the fake score. We apply gradient norm clipping with a value of 1.0 for both the generator and fake score. We use batch size 64. We set the CFG as 8. Since SDXL has 2.7B parameters, fine-tuning it at 1024 resolution is computationally expensive. We first fine-tune for 1k iterations at 512 resolution, then fine-tune for another 1k iterations at 1024 resolution. The fake score is initialized from the pre-trained SDXL in both stages.

\paragraph{PixArt-$\alpha$} We adopt a constant learning rate of 2e-6 for the generator and 2e-5 for the fake score. We set the CFG as 3.5. We apply gradient norm clipping with a value of 1.0 for both the generator and fake score. We use batch size 32. The training can be done within 500 iterations.

\section{Additional Experiments}
\label{app:add_exp}

\begin{figure}
    \centering
    \includegraphics[width=1\linewidth]{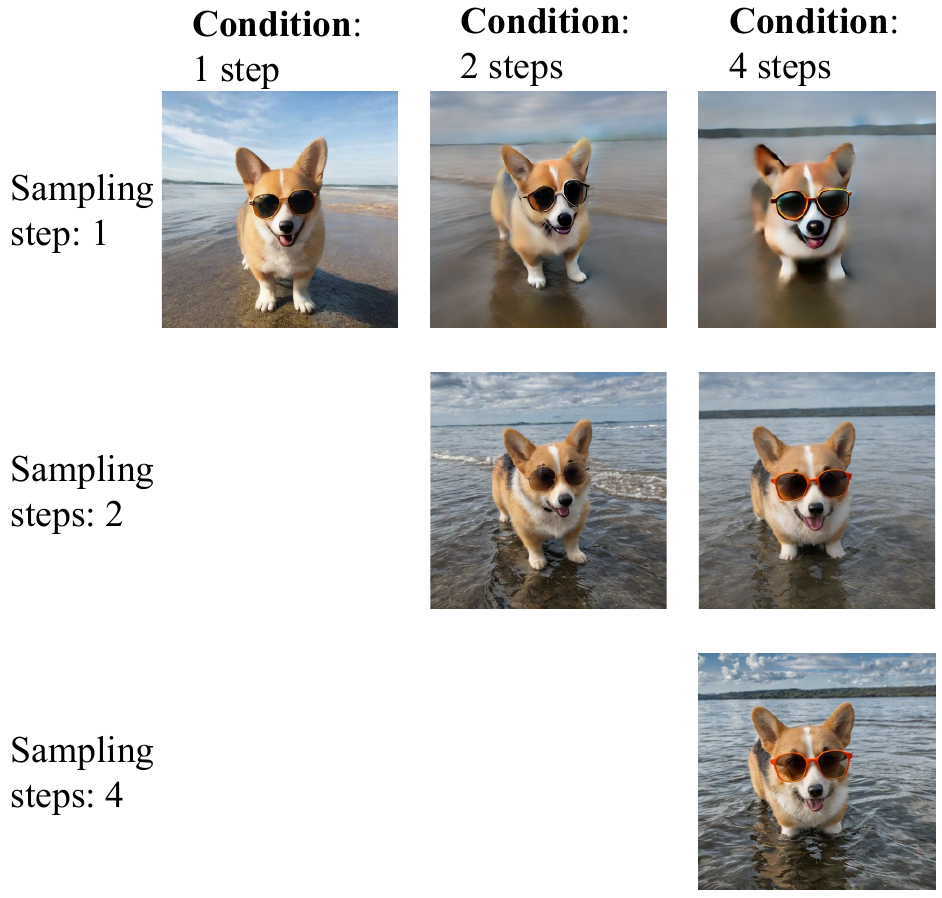}
    \caption{Visual samples of varying the condition steps and sampling steps. The prompt is ``A corgi with sunglasses, traveling in the sea""}
    \label{fig:flexible}
\end{figure}

\paragraph{The Effect of Varying Sampling Steps.} We investigate the impact of different sampling and conditioning configurations, with results visualized in \cref{fig:flexible}. The results show that TDM-unify exhibits systematic behavioral variations across different combinations of conditional and sampling steps. Notably, the model demonstrates an intrinsic understanding of the underlying ODE trajectory, adaptively positioning itself at appropriate points along this path given the specified condition steps. 

\section{Ablation Studies}
\label{app:add_ablation}

\subsection{Ablation Details}
We use the same hyperparameters and training iterations for all variants, with differences only in the ablating components.

\paragraph{The formulation of using GANs during distillation} Following the previous work~\cite{sdxllight,yoso}, we use latent discriminators, with the backbone based on the UNet encoder from SD-v1.5. In particular, we perform the following loss for learning generator:
\begin{equation}
\small
\begin{aligned}
    \E_K
    \sum_{i=0}^{K-1}\sum_{\tau = t^{K}_i}^{t^{K}_{i+1}} \{
    & \lambda_\tau\mathrm{KL}(p_{\theta,\tau|t^{K}_i}(\rvx_\tau|K)  || p_{\phi,\tau}(\rvx_{\tau})) + \\
    & \lambda_\tau \E_{p_{\theta,\tau|t^{K}_i}(\rvx_\tau|K)}\mathrm{ADVLoss} (\rvx_\tau, \tau, K)\}.
\end{aligned}
\end{equation}
The $\mathrm{ADVLoss}$ is the adversarial loss. Note that for a fair comparison, we inject the desired sampling steps $K$ into GAN's discriminator too.

\paragraph{Details in implementing original loss} We use normalization proposed in DMD~\cite{yin2023one}, while we do not use the normalization in our proposed surrogate loss.

\paragraph{Details in implementing DMD2} Following the original DMD2 paper~\cite{dmd2}, we update the fake score 10 times per iteration. Other hyper-parameters remain consistent with our configuration.

\begin{table}[!t]
    \centering
    \caption{Comparison on HPS across variants in 4-step generation based on SD-v1.5.}
    \vspace{-2mm}
    \label{tab:effect_ode}
    \resizebox{1\linewidth}{!}{
    \begin{tabular}{ccccc}
    \toprule
     Train-DDIM & Train-DPMSolver & Test-DDIM & Test-DPMSolver & HPS$\uparrow$\\
     \midrule
    \checkmark & & \checkmark & & 31.04 \\
    \checkmark & &  & \checkmark & 31.35 \\
    & \checkmark &  \checkmark &  & 30.86 \\
     & \checkmark &  & \checkmark & 31.30 \\
    \bottomrule
    \end{tabular}
    }
    % \vspace{-5mm}
\end{table}

\begin{table}[!t]
    \centering
    \caption{Comparison on HPS across variants in 4-step generation based on SD-v1.5.}
    \vspace{-2mm}
    \label{tab:match_noisy}
    \begin{tabular}{lc}
    \toprule
    Method & HPS$\uparrow$ \\
    \midrule
    TDM (Matching noisy samples $\rvx_{t_i}$) & 31.35 \\
    Matching clean samples $\hat \rvx_{t_i}$ & 24.63 \\
    \bottomrule
    \end{tabular}
    % \vspace{-5mm}
\end{table}

\begin{table}[!t]
    \centering
    \caption{The effect of using more expensive Fisher Divergence in 4-step generation.}
    \vspace{-2mm}
    \label{tab:fisher}
    \begin{tabular}{lc}
    \toprule
    Method & HPS$\uparrow$ \\
    \midrule
    TDM  & 31.35 \\
    TDM w/ Fisher & \textbf{31.70} \\
    \bottomrule
    \end{tabular}
\end{table}

\subsection{Additional Ablation}
To gain a more comprehensive understanding of our proposed methods, We conducted additional ablation studies in this section.

\begin{figure}[t]
    \centering
    \subfloat[DMD2]{
        \includegraphics[width=0.95\linewidth]{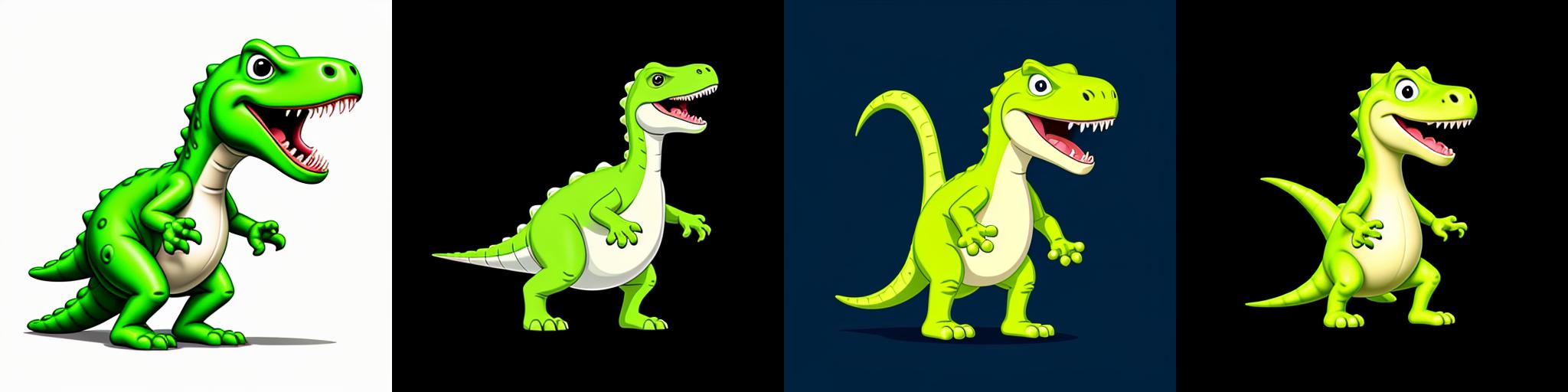}
    }\\
    \subfloat[Ours w/o Importance Sampling]{
        \includegraphics[width=0.95\linewidth]{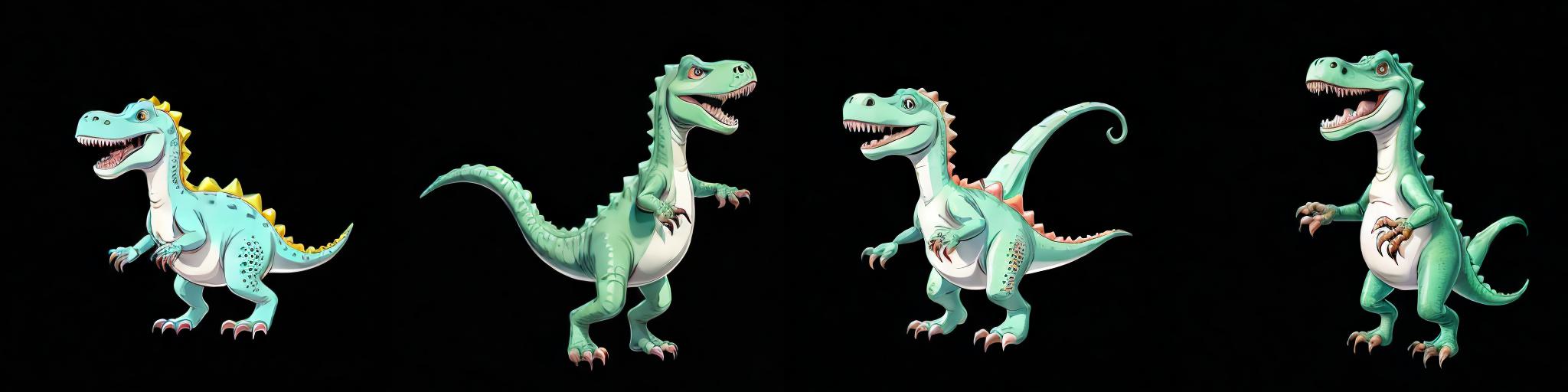}
    }\\
    \subfloat[Ours]{
        \includegraphics[width=0.95\linewidth]{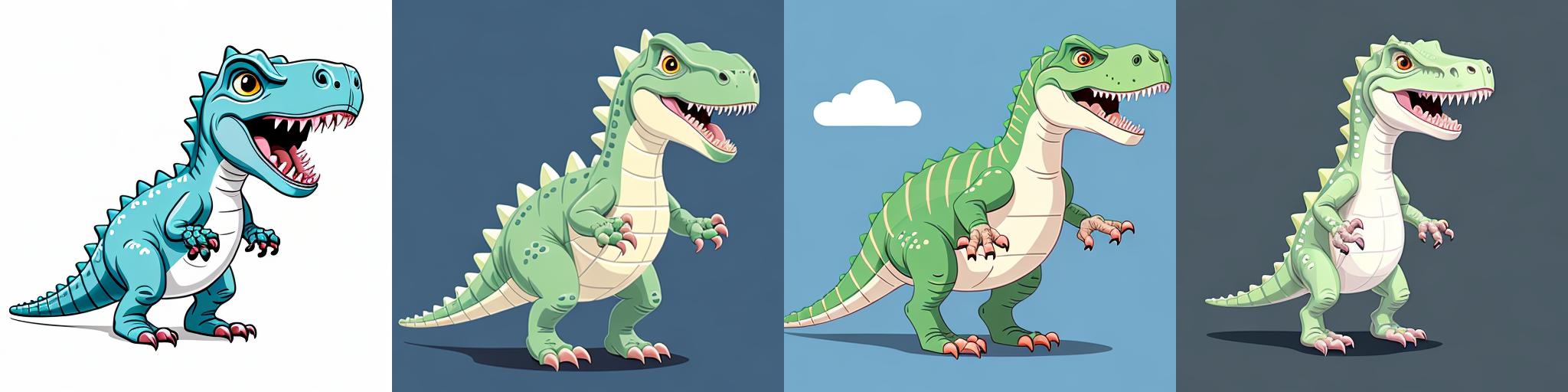}
    }
    \caption{Comparison on Mode Cover in 4-step generation based on SD-v1.5. It is clear that our method has better mode cover and image quality. The prompt is ``A cute dinosaur, cartoon style"}
    \label{fig:mode} 
\end{figure}

\begin{figure}[t]
    \centering
    \subfloat[matching clean samples]{
        \includegraphics[width=0.95\linewidth]{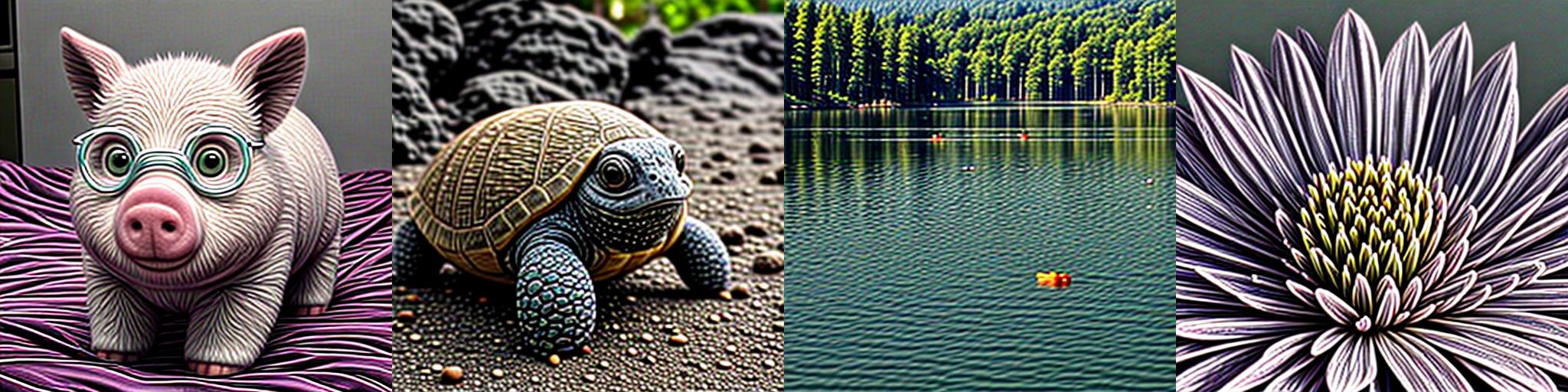}
    }\\
    \subfloat[Ours (matching noisy samples)]{
        \includegraphics[width=0.95\linewidth]{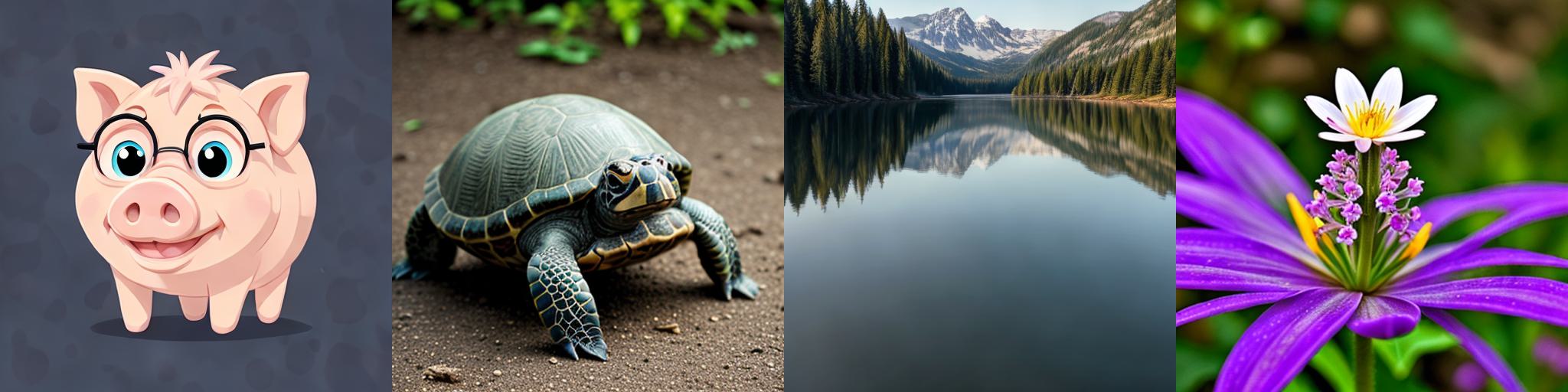}
    }
    \caption{
    Comparison on the compatibility with deterministic samplers in the 4-step generation on SD-v1.5.
    It is clear that our method (matching noisy samples) has better visual quality.}
    \label{fig:match_comp} 
\end{figure}

\paragraph{Comparison on Mode Coverage.} We found that using importance sampling for learning the fake score led to improved performance (\cref{tab:ablation}) and better mode coverage. As shown in \cref{fig:mode}, our method demonstrates notably superior image quality and mode coverage. This improvement may be attributed to the fact that the fake score cannot accurately track the student distribution without using importance sampling. We additionally compare to the concurrent work DMD2~\cite{dmd2}. We found that DMD2's generated results also suffer from mode collapse, while its impact is somewhat less severe compared to the variant without importance sampling. This may be due to DMD2 trains the fake score multiple times at each iteration, resulting in a more accurate fake score at the cost of slow training.

\paragraph{Effect of Different ODE Solvers.} In the experiments presented in the main body, we use DDIM~\cite{ddim} as the ODE solver during training and DPMsolver~\cite{lu2023dpm} during inference. Here we ablate the choice of ODE solver, with results shown in \cref{tab:effect_ode}. We find that using DDIM versus DPM during training yields similar performance, which may be attributed to two reasons: 1) regardless of whether DDIM or DPM is used during training, we can utilize DPM sampling at test time to improve performance; 2) our training only backpropagates through one ODE step, preventing higher-order ODE solvers like DPMSolver from benefiting from higher-order information in the training.

\paragraph{Matching noisy samples $\rvx_{t_i}$ v.s. clean samples $\hat \rvx_{t_i}$}  A core design of our method is to align noisy samples $\rvx_{t_i}$ predicted by the model with the target diffusion, rather than the clean samples $\hat \rvx_{t_i}$ predicted by the model. This design makes the support of deterministic sampling possible. We conduct experiments on matching clean samples, the results are shown in \cref{tab:match_noisy} and \cref{fig:match_comp}. It is clear that our method has a better performance, while matching clean samples deliver a poor deterministic sampling with notable artifacts.

\paragraph{Flexibility for using different distribution divergence} The proposed TDM has the flexibility for using different distribution divergence instead of reverse KL divergence. In particular, the performance of TDM can be further improved by more expensive Fisher divergence (\cref{tab:fisher}).

\section{User Study Details}
\label{app:user}

\begin{figure}
    \centering
    \includegraphics[width=1\linewidth]{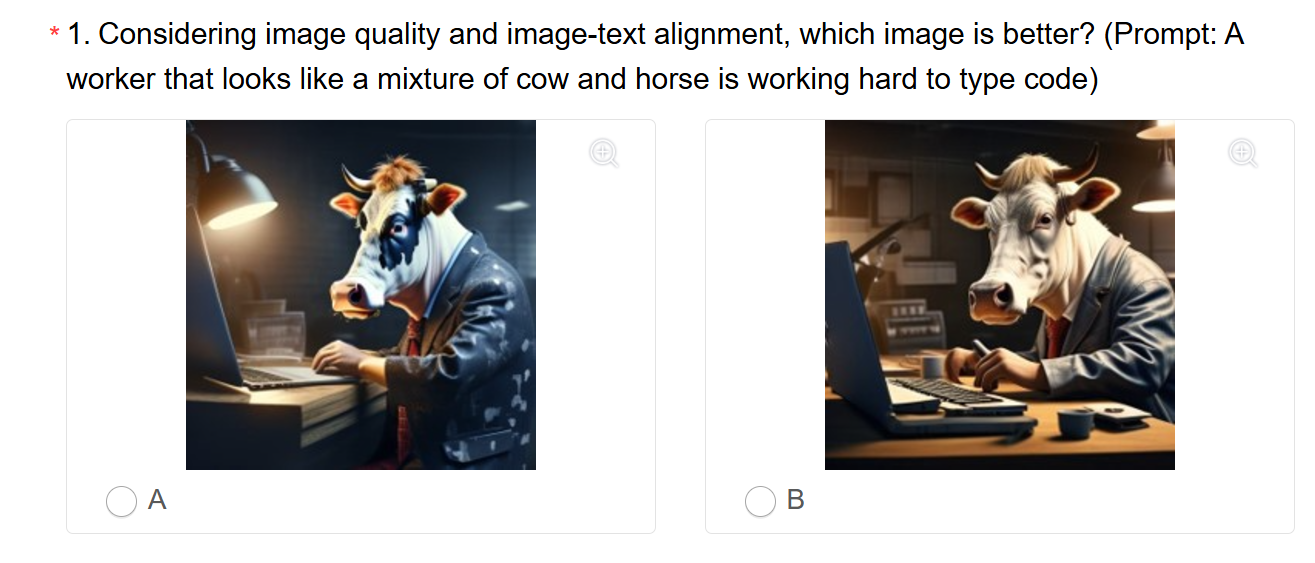}
    \caption{An example of the evaluation question for our user study.}
    \label{fig:user_demo}
\end{figure}

We conducted user research by presenting users with two anonymous images generated by different models and asking them to select the sample with higher image quality and better prompt alignment. We randomly selected 20 prompts for image generation. 
Each image was manually verified to ensure the absence of inappropriate or dangerous content. An example of an evaluation question is shown in \cref{fig:user_demo}. In total, we collected approximately 40 user responses.

\section{Additional Qualitative Results}
\label{app:add_imgs}

\begin{figure*}
    \centering
    \includegraphics[width=1\linewidth]{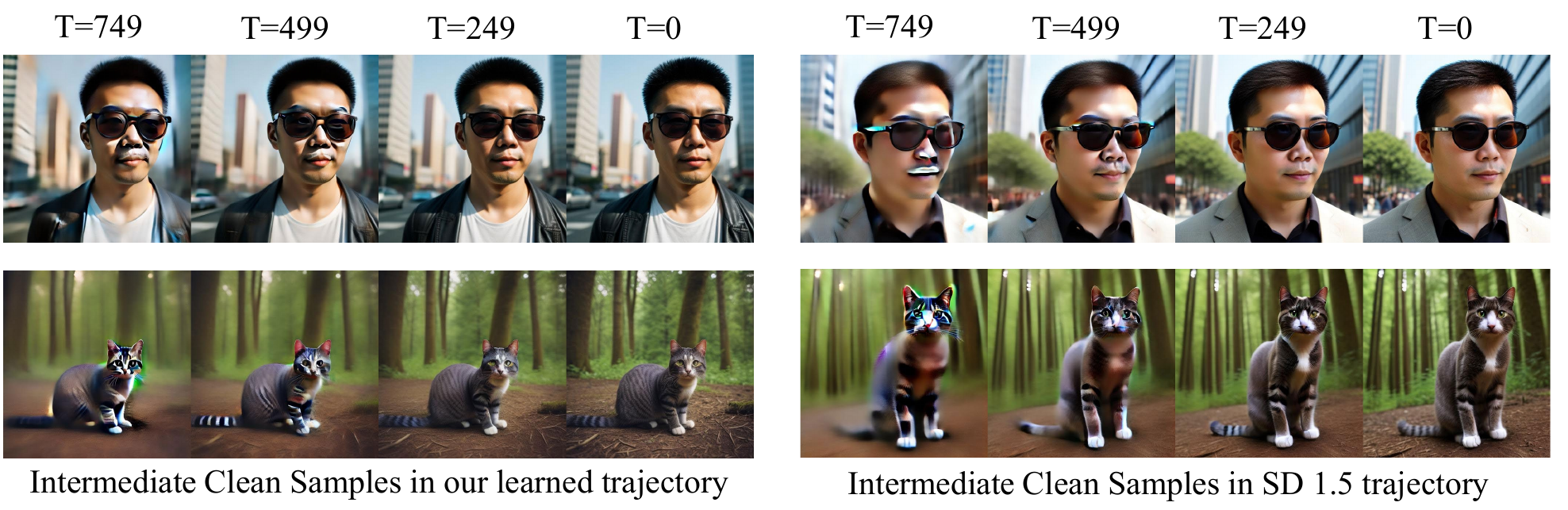}
    \caption{Additional visualization of ODE trajectory with clean samples at different timesteps. It is clear that our method suffers less from the CFG artifact and has better visual quality.}
    \label{fig:more_traj}
\end{figure*}

We present the additional visualization of ODE trajectory with clean samples at different timesteps in \cref{fig:more_traj}.

\begin{figure*}
    \centering
    \includegraphics[width=1\linewidth]{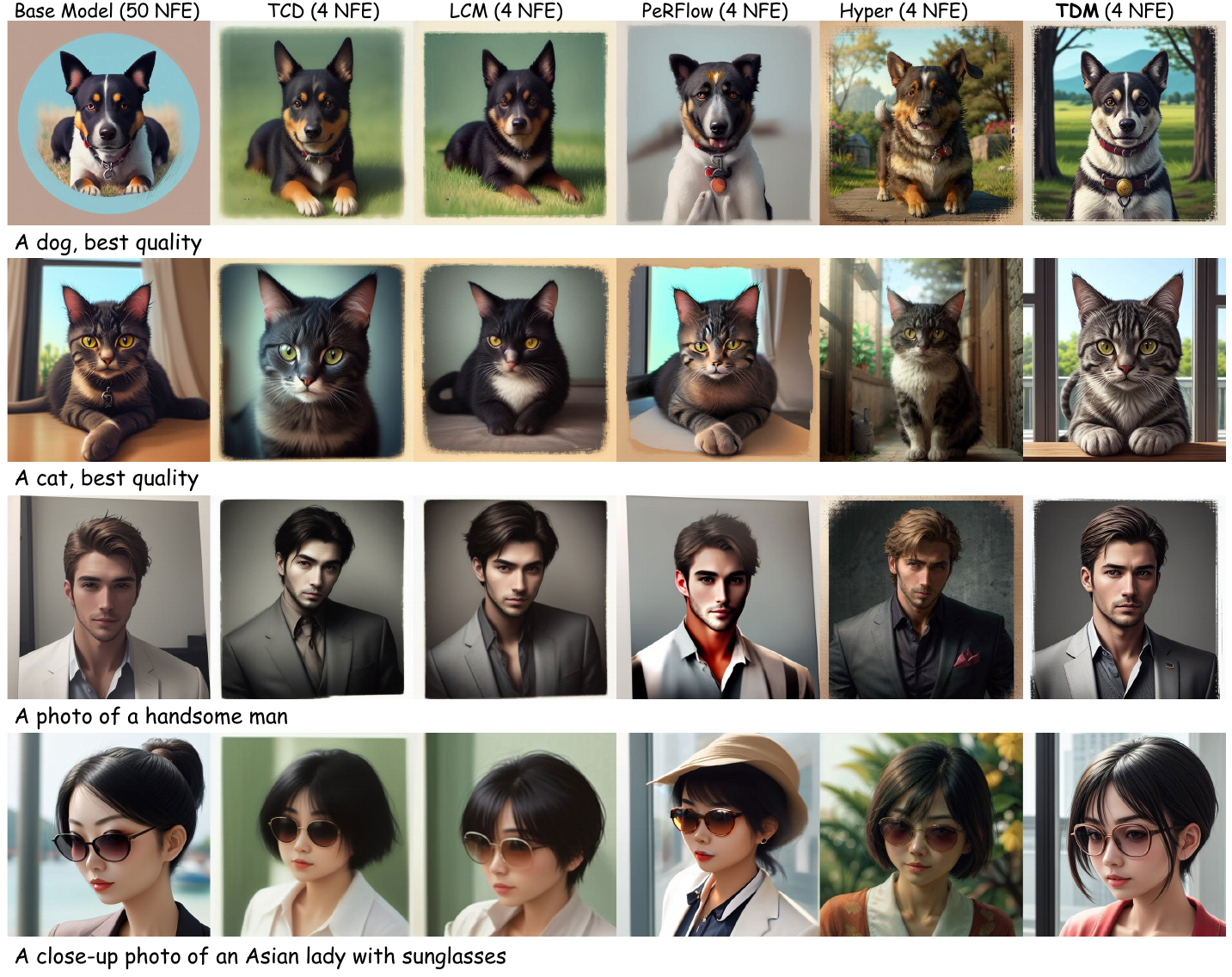}
    \caption{Additional Samples of integrating LoRA into unseen customized models - dreamshaper.}
    \label{fig:style1}
\end{figure*}

\begin{figure*}
    \centering
    \includegraphics[width=1\linewidth]{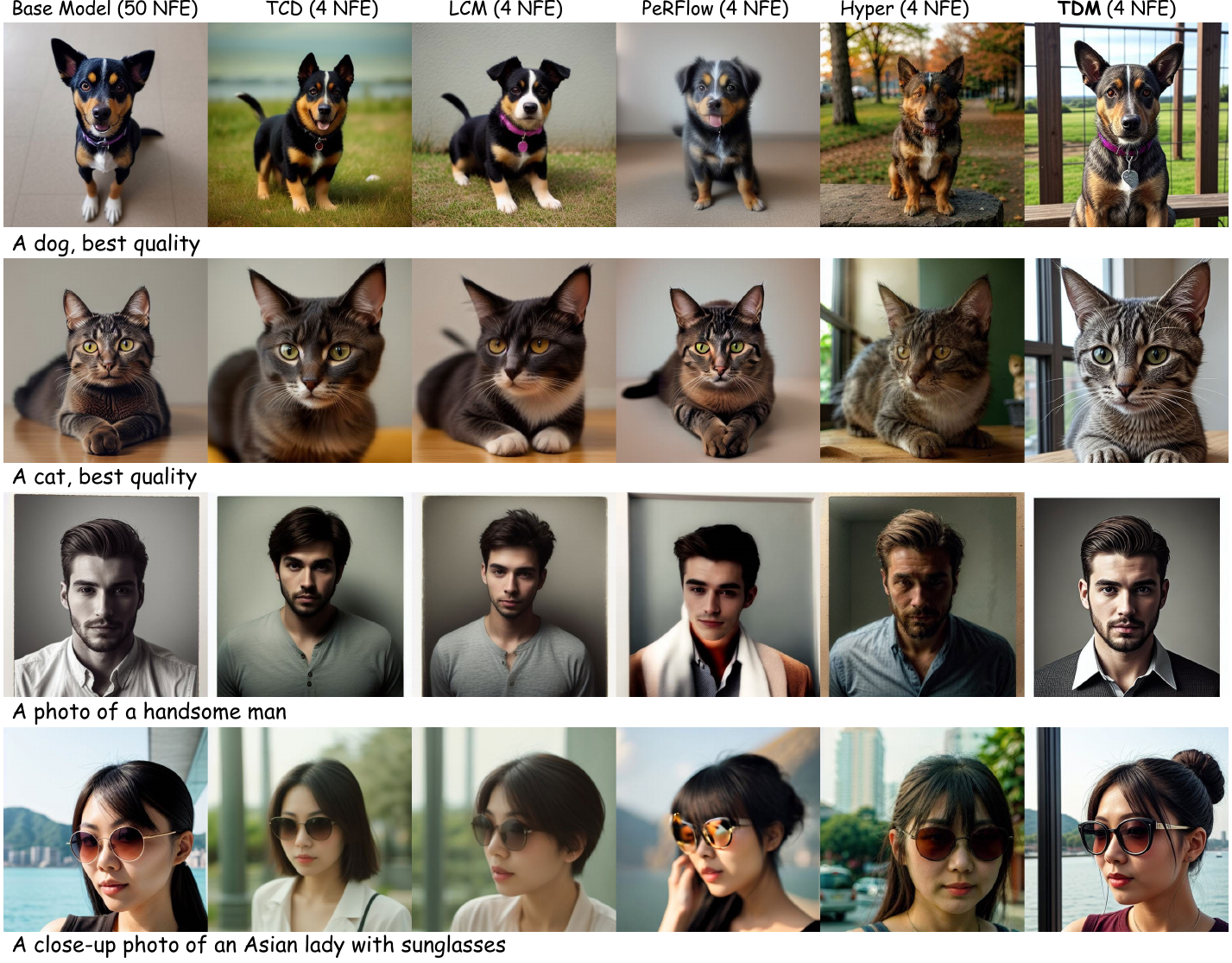}
    \caption{Additional Samples of integrating LoRA into unseen customized models - realisticvision.}
    \label{fig:style2}
\end{figure*}

We present the visual samples of interesting LoRA into unseen customized models in \cref{fig:style1,fig:style2}. It can be seen that compared to the competing baseline, our method shows better visual quality and better style preservation.

% We present the additional visual samples of TDM on the SDXL backbone in \cref{fig:add_sdxl}.

\label{app:addtional_visual}

\end{document}